# Epistemic Memory: A Validity Layer for Self-Maintaining Intelligent Systems

Pin-Han Ho[1] , Limei Peng[1,2] , Yiming Miao[1] , Yan Jiao[1]

[1] Shenzhen Institute for Advanced Study,

University of Electronic Science and Technology of China;

[2] Kyungpook National University

Correspondence: Pin-Han Ho, pinhanho71@gmail.com

Artificial intelligence has progressively expanded its mechanisms for storing, representing, retrieving, and adapting knowledge. However, existing AI memory paradigms primarily focus on preserving information content, while the validity conditions under which knowledge remains applicable are often treated implicitly. This gap gives rise to a phenomenon we term *semantic coordinate drift*: when an embodied agent moves, changes sensors, or encounters novel environmental configurations, previously valid semantic distinctions may become unsupported while new distinctions emerge, yet the agent's internal representation remains fixed. This paper suggests that the evolution toward self-maintaining intelligent systems motivates an emerging memory paradigm—epistemic memory. Rather than replacing existing memory mechanisms, epistemic memory provides a validity-maintenance layer that governs when stored knowledge remains applicable. Here, self-maintaining intelligence refers to the capability of preserving and reorganizing the validity of knowledge representations under changing observation conditions. Epistemic memory explicitly tracks not only what is known, but also the boundary of knowability under current observation conditions. We formalize the dynamic epistemic quotient, an observation-induced equivalence structure over hypotheses that evolves with an agent's interaction history and sensing capabilities. We derive a pairwise incompatibility lower bound showing that any fixed semantic representation must incur unavoidable error across changing epistemic boundaries. We further identify a failure mode under symmetric crossing quotients, where overlap-based class-level transport becomes independent of pre-transition belief, motivating the preservation of within-class hypothesis provenance. We introduce Observable Belief Memory (OBM) as a constructive epistemic governance architecture, with state $M_t = (Q_t, b_t, \rho_t)$, combining the current epistemic quotient, belief over quotient classes, and within-class provenance through a posterior-consistent update. Controlled experiments demonstrate that explicit epistemic tracking improves robustness under changing observation conditions, particularly when class-level correspondence becomes insufficient. These results suggest that self-maintaining intelligent systems may benefit from reasoning not only about hidden states, but also about the evolving validity of the representations through which those states become knowable.

# 1 Introduction: The Missing Validity Layer in Artificial Intelligence

The evolution of artificial intelligence can be understood through the lens of memory. Early reactive systems maintained immediate sensory states and responded through predefined mappings. Deep learning introduced representational memory, where embedding spaces capture reusable semantic structures from data Bengio et al. (2013). Foundation models further expanded this paradigm into parametric memory, storing large-scale knowledge within model parameters Bommasani et al. (2021). Despite these advances, existing AI memory paradigms primarily focus on storing, representing, and retrieving information. A critical dimension remains largely implicit: *maintaining the validity conditions under which stored knowledge remains justified*.

Consider an embodied agent navigating an indoor environment. Under normal illumination, visual observations may distinguish objects through color, texture, and shape. After environmental changes, however, only depth or thermal signatures may remain reliable. The semantic categories learned during previous observations—such as “red ball” or “blue box”—may remain encoded in the agent’s latent representations, while the current observation process no longer supports these distinctions. This is the essence of *semantic coordinate drift*: the agent retains what it has learned, but it lacks an explicit mechanism to determine whether that knowledge remains justified under the current sensing conditions. This phenomenon is illustrated in Fig. 2. When observation conditions change, the semantic distinctions preserved by a fixed representation may no longer correspond to the currently observable epistemic structure. As a result, previously valid knowledge representations may become incompatible with the new observation-induced boundary.

This limitation extends beyond embodied robotics. A large language model may encode the fact that “Paris is the capital of France,” but current architectures generally do not explicitly represent the contextual, observational, and evidential conditions under which such knowledge should remain applicable. Similarly, world models Ha and Schmidhuber (2018); Hafner et al. (2019, 2020, 2023) learn predictive latent dynamics but typically assume that learned latent distinctions preserve consistent meaning across changing conditions. Continual learning systems Kirkpatrick et al. (2017); Parisi et al. (2019) address catastrophic forgetting French (1999), yet primarily focus on preserving previously acquired knowledge rather than determining when such knowledge should be reorganized, weakened, or retired.

These challenges motivate a broader vision of *self-maintaining intelligent systems*. In this work, *self-maintaining intelligence* refers specifically to the capability of an intelligent system to continuously preserve and reorganize the validity of its own knowledge representations under changing observation conditions, environmental dynamics, and available evidence. It does not refer to autonomous self-repair, unrestricted self-modification, or complete independence from external supervision. Instead, self-maintaining intelligence describes the ability of an agent to monitor whether its knowledge remains applicable, update the boundaries of what can be inferred, and adapt its internal organization accordingly.

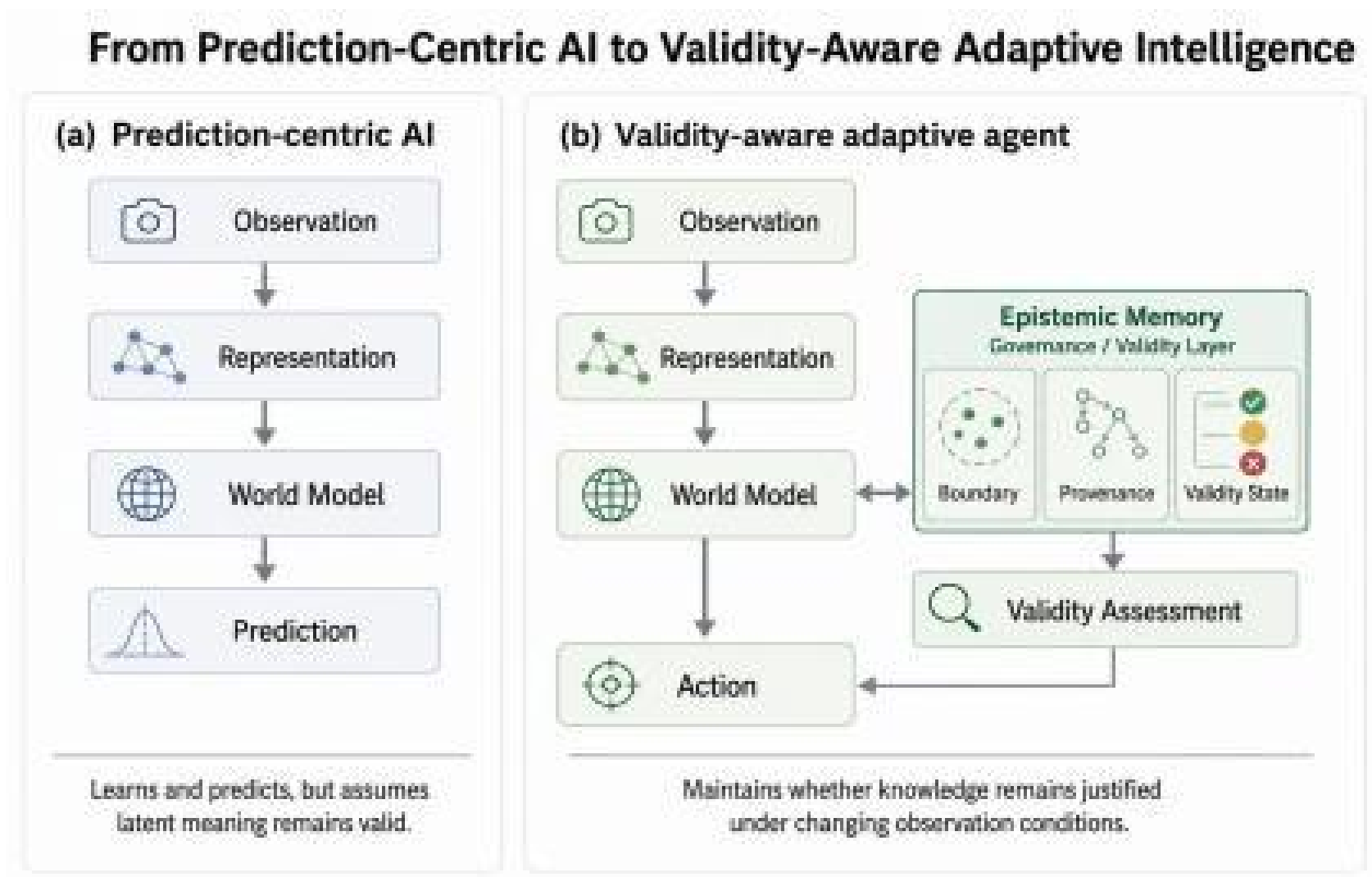

**Fig. 1** From prediction-centric AI to validity-aware adaptive intelligence. Existing AI systems optimize observation, representation, world modeling, and prediction under assumed stable semantic validity. Self-maintaining intelligence introduces epistemic memory as a governance layer that tracks boundary, provenance, and validity state.

We argue that achieving such capability requires a new direction in AI memory architectures: *epistemic memory*. Unlike conventional memory mechanisms that primarily store information content, epistemic memory maintains three interlocking components:

1. **Knowledge**: what the agent believes about the world;
2. **Boundary**: which distinctions are currently justified by available observations and evidence;
3. **Provenance**: the distribution of hypotheses within each epistemic class, enabling future knowledge reorganization.

Therefore, epistemic memory should not be interpreted as another storage mechanism or replacement for neural memory architectures. Instead, it serves as an epistemic governance layer that determines when stored knowledge remains meaningful, under what conditions it should be trusted, and how it should be reorganized when the agent's observation boundary changes.

This transition is illustrated in Fig. 1, where conventional prediction-centric AI systems are contrasted with validity-aware adaptive systems that incorporate epistemic memory as an additional layer for maintaining knowledge validity under changing observation conditions.

***Why now: From static models to adaptive intelligent agents.***

The need for epistemic memory emerges from a fundamental transition in artificial intelligence: the shift from static predictive models toward adaptive intelligent agents operating in dynamic environments. Traditional AI systems are often evaluated under relatively fixed assumptions, where the observation process, task definition, and representation space remain stable after training. Under these conditions, storing knowledge and improving predictive accuracy are sufficient objectives.

The emergence of foundation models and autonomous agents changes the role of memory in artificial intelligence. Earlier AI systems primarily performed bounded prediction tasks, where learned representations were evaluated under relatively fixed distributions. In contrast, modern agents continuously interact with dynamic environments, accumulate experience, retrieve prior knowledge, and modify their own behavior. Large-scale models are increasingly deployed beyond their original training distributions, where retrieved knowledge, learned representations, and generated decisions may encounter contexts whose validity conditions were not explicitly modeled during training. Embodied agents continuously modify their viewpoints, sensing configurations, and interaction histories; autonomous systems encounter novel environments; and foundation models are deployed across diverse contexts beyond their original training distributions.

Under these conditions, the central challenge is no longer only how to store knowledge, but how to determine whether previously acquired knowledge remains valid under current conditions. A system that can remember information but cannot evaluate the validity of its own representations may continue to retrieve and apply knowledge beyond the conditions that justify it.

Therefore, as AI evolves from static models toward adaptive, self-maintaining agents, the ability to maintain the relationship between knowledge and its validity conditions becomes an increasingly important architectural capability. This motivates the development of epistemic memory as a mechanism for explicitly tracking the relationship between knowledge, observation conditions, and the evolving boundary of knowability.

This paradigm shift demands a formal framework. We introduce the *dynamic epistemic quotient*, an observation-induced equivalence structure over hypotheses that evolves with an agent's interaction history and available sensing capabilities. The key departure from existing abstraction frameworks is not merely the introduction of another equivalence relation; rather, the equivalence structure itself becomes a dynamic state variable, evolving with the agent's interaction history and sensing capability.

### Contributions.

This work makes three principal contributions:

1. **Conceptual shift:** We identify the missing validity layer in AI memory and articulate epistemic memory as an emerging paradigm that extends existing AI memory mechanisms with explicit validity maintenance. We introduce *semantic coordinate drift* as a previously unnamed phenomenon arising from the mismatch between fixed semantic coordinates and dynamically evolving epistemic boundaries.

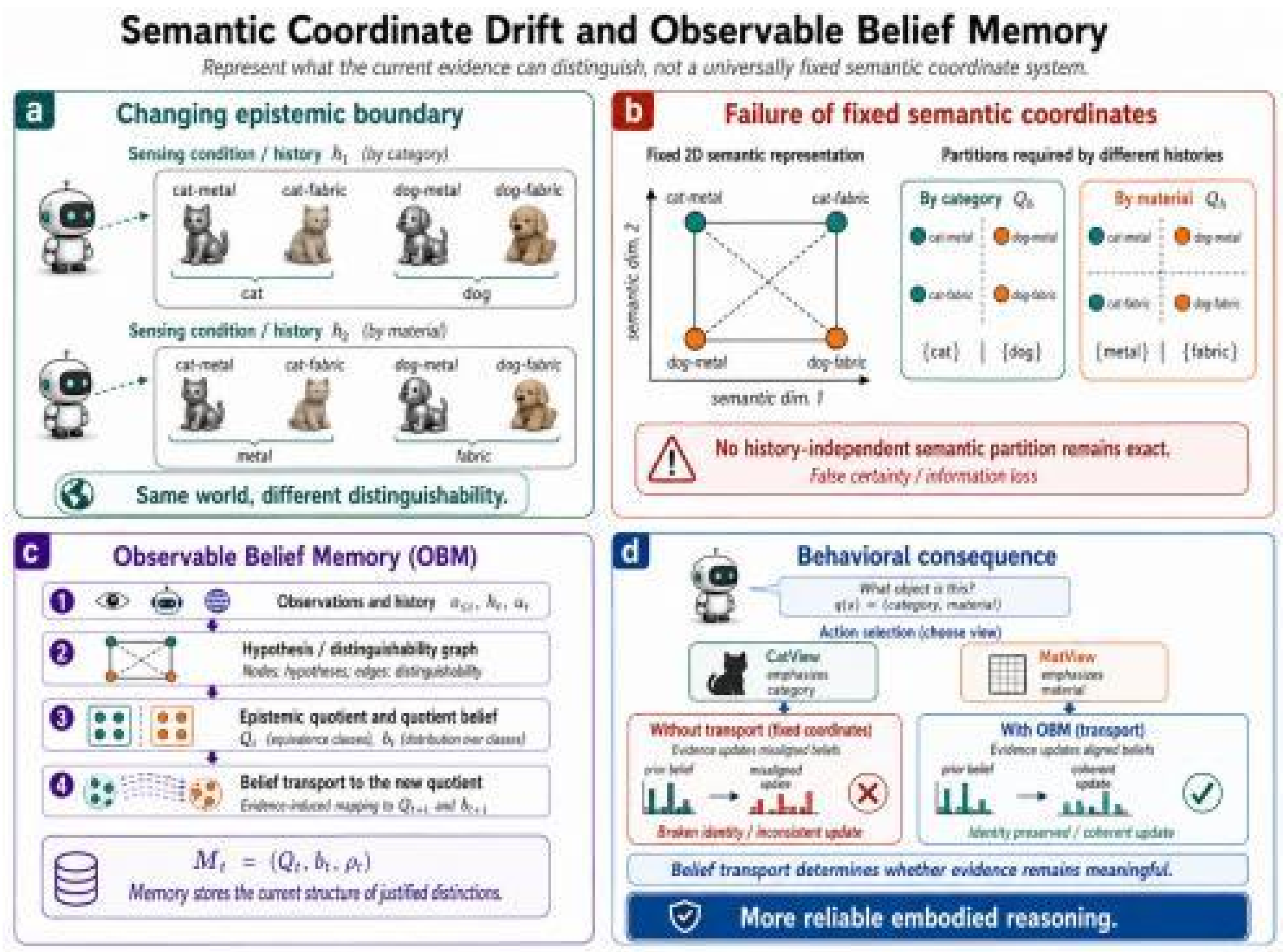

**Fig. 2 Semantic coordinate drift and validity-aware memory organization.** (a) A fixed semantic representation assumes that learned semantic coordinates remain valid across observation conditions. When the epistemic quotient changes, the same representation may become incompatible with the newly observable distinctions. (b) Epistemic memory addresses this mismatch by maintaining the evolving epistemic boundary together with belief and provenance information, enabling knowledge reorganization across changing observation conditions.

2. **Formal foundation:** We formalize the epistemic quotient as a dynamic equivalence structure over hypotheses induced by observation history and available sensing capability. We derive a pairwise incompatibility lower bound showing that any fixed representation must incur unavoidable error across distinct epistemic quotients, and identify a failure mode under symmetric crossing quotients that motivates within-class provenance preservation.
3. **Architectural implication:** We introduce Observable Belief Memory (OBM) as a constructive epistemic governance layer with state $M_t = (Q_t, b_t, \rho_t)$, combining the current epistemic quotient, belief distribution, and within-class provenance through a posterior-consistent update. We provide empirical demonstrations across controlled and embodied settings illustrating recoverability, boundary evolution, and the advantages of explicit epistemic tracking.

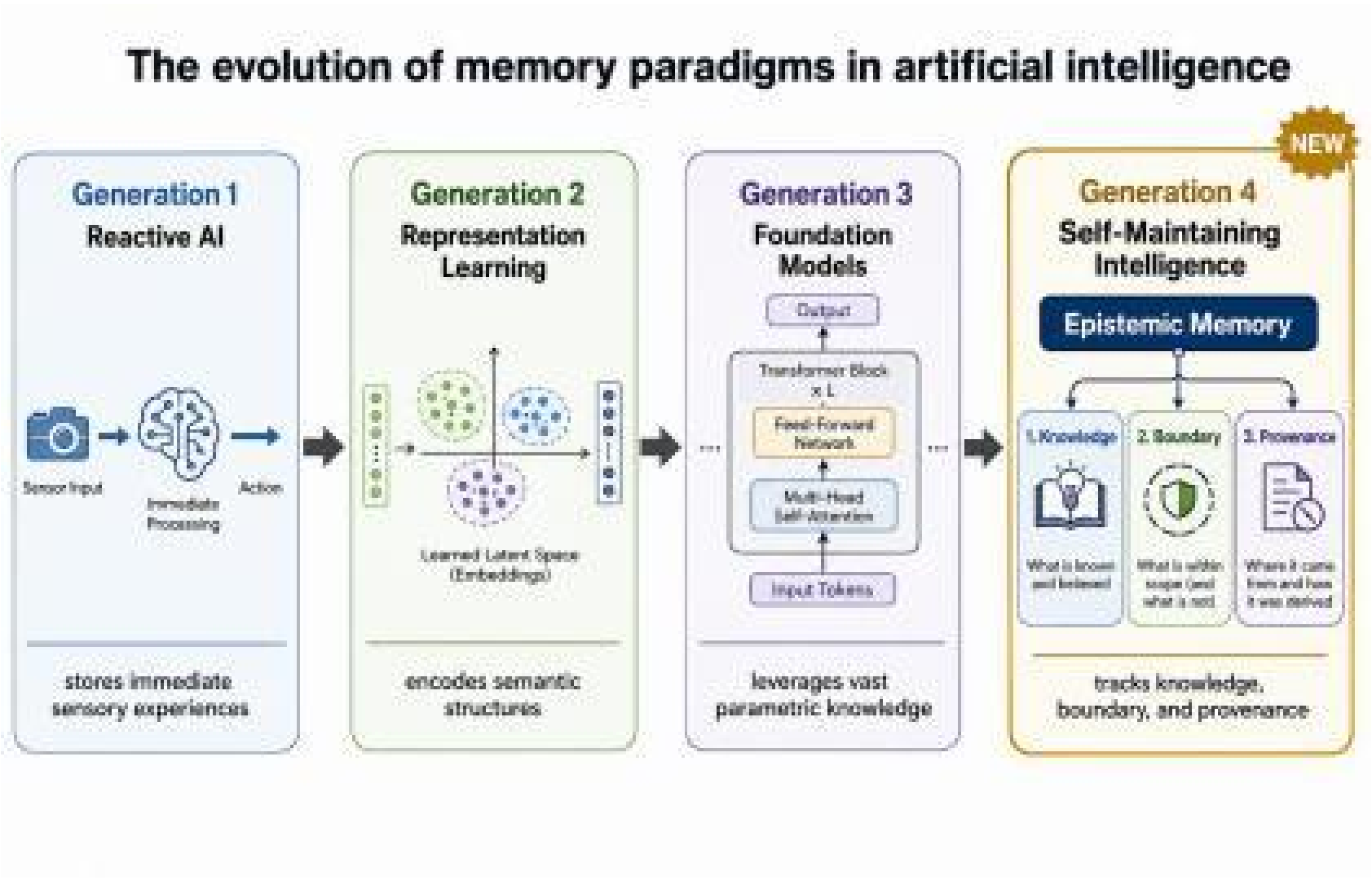

**Fig. 3 The evolution of memory paradigms in artificial intelligence.** Generation 1: Reactive AI stores immediate sensory experiences. Generation 2: Representation learning encodes semantic structures. Generation 3: Foundation models leverage vast parametric knowledge. Emerging direction: Self-maintaining intelligence extends existing AI memory mechanisms with explicit validity maintenance through epistemic memory.

## 2 Evolution of Memory Paradigms in Artificial Intelligence

Fig.3 summarizes this historical transition from storage-oriented to validity-aware memory mechanisms. The evolution of artificial intelligence can be viewed through the progressive transformation of memory mechanisms. Early AI systems focused on storing explicit knowledge, whereas modern systems increasingly rely on distributed representations, predictive latent models, and adaptive agent memories. Despite these advances, a common assumption remains across most existing paradigms: once knowledge is acquired, the semantic conditions under which that knowledge remains valid are implicitly assumed to remain stable.

This section reviews the evolution of AI memory paradigms from four perspectives: representation, prediction, adaptation, and reliability. We show that although existing approaches progressively improve the ability to store, use, and evaluate knowledge, they generally lack an explicit mechanism for maintaining knowledge validity under changing observation conditions.

### 2.1 Cognitive Memory Foundations

Cognitive science provides a conceptual foundation for distinguishing different forms of memory. Human memory is commonly divided into episodic memory, which preserves specific experiences and events, and semantic memory, which captures generalized knowledge structures independent of particular episodes Tulving (1972, 1985). This distinction reflects an important insight: intelligent systems must not only store information but also organize it according to different functional roles.

Metacognition, the ability to monitor and regulate one's own cognitive processes, further highlights the importance of awareness of knowledge states Flavell (1979); Nelson and Narens (1990). Human cognition does not simply represent the world; it also maintains awareness of the reliability, limitations, and applicability of its own knowledge. This capability enables adaptive behavior under changing conditions, where agents must assess when previously acquired knowledge remains relevant.

However, while cognitive science has extensively studied awareness of knowledge states, existing AI implementations rarely formalize the dynamic validity boundary of representations under changing observation conditions. Cognitive architectures such as ACT-R and SOAR have incorporated mechanisms for memory, reasoning, and goal management Anderson et al. (2004); Newell (1990), but they generally assume stable representational validity.

## 2.2 From Explicit Knowledge Storage to Learned Representations

Classical artificial intelligence viewed memory primarily as explicit knowledge storage. Symbolic systems maintained facts, rules, and logical relations that could be queried and manipulated through reasoning procedures. These approaches provided explicit and interpretable representations of knowledge. However, they generally assumed that stored knowledge remained valid once acquired, while changes in applicability were handled through external updates or human intervention.

Deep learning introduced a different memory paradigm based on learned representations. Instead of explicitly storing symbolic facts, neural networks encode knowledge in distributed latent representations and model parameters Bengio et al. (2013). Representation learning, variational approaches, and self-supervised learning demonstrated that useful semantic structures can emerge from data-driven optimization Kingma and Welling (2014); Chen et al. (2020); Grill et al. (2020). Foundation models further extended this paradigm by embedding large-scale knowledge into billions of parameters, enabling broad generalization across tasks Devlin et al. (2019); Bommasani et al. (2021).

However, both explicit and neural representation paradigms share an implicit assumption: the semantic organization of knowledge remains sufficiently stable after learning. A symbolic concept, latent dimension, or embedding direction is expected to preserve its meaning across contexts. When observation conditions change and previously meaningful distinctions become invalid, these systems generally lack an explicit mechanism for recognizing that their internal semantic structure requires revision. Representation learning generally focuses on discovering useful coordinates rather than maintaining when those coordinates remain valid.

## 2.3 From Predictive Models to Adaptive Agents

Beyond storing representations, modern AI systems increasingly rely on memory for prediction, planning, and interaction. World models learn compact latent dynamics that enable agents to simulate future states and perform planning in imagined environments Ha and Schmidhuber (2018); Hafner et al. (2019, 2020, 2023); Schrittwieser et al. (2020). Predictive state representations provide another perspective by representing environments through predictive information rather than explicit latent states Littman et al. (2001); Singh et al. (2004).

State abstraction and bisimulation-based approaches further investigate when different states can be considered equivalent while preserving decision-relevant information Givan et al. (2003); Li et al. (2006); Ravindran and Barto (2004). These approaches provide important foundations for understanding representation compression and abstraction in sequential decision-making.

Embodied intelligence extends these ideas by emphasizing that knowledge emerges through interaction between an agent and its environment Brooks (1991); Pfeifer and Bongard (2007); Clark (1998). Active perception enables agents to select observations and actions that improve information acquisition Bajcsy (1988); Aloimonos et al. (1988). Information-theoretic sensing further formalizes observation selection through measures of expected information gain Lindley (1956); MacKay (2003); Krause and Guestrin (2008).

Recent autonomous agents combine these ideas with episodic memory, retrieval mechanisms, and planning memories. Memory-augmented neural networks provide external storage mechanisms for sequential reasoning and decision-making Graves et al. (2014); Weston et al. (2015); Santoro et al. (2016); Rae et al. (2016). Generative agents simulate human-like behavior through memory streams, reflection, and planning Park et al. (2023). MemGPT introduces operating system-inspired memory management for large language models Packer et al. (2023). However, predictive models and agent memory systems generally assume that stored representations remain applicable across changing environments. They can update predictions or retrieve relevant experiences, but they do not explicitly determine when the semantic distinctions underlying those memories should be reorganized.

## 2.4 From Knowledge Preservation to Reliability Assessment

As AI systems became increasingly deployed in dynamic environments, researchers developed mechanisms for maintaining reliability over time. Continual learning addresses the challenge of catastrophic forgetting, where neural networks overwrite previously acquired knowledge when learning new tasks French (1999); Kirkpatrick et al. (2017); Zenke et al. (2017); Parisi et al. (2019). These approaches focus on preserving learned parameters and preventing destructive updates.

A related research direction investigates uncertainty estimation and confidence reliability. Bayesian approaches distinguish different sources of predictive uncertainty and provide principled mechanisms for estimating model confidence Neal (1996); MacKay (1992); Kendall and Gal (2017). Deep ensembles and Bayesian approximations further improve uncertainty estimation in modern neural networks Gal and Ghahramani (2016); Lakshminarayanan et al. (2017).

Trustworthy AI extends reliability considerations toward robustness, interpretability, accountability, and safe deployment. These efforts provide essential mechanisms for evaluating whether AI outputs can be trusted under various conditions Arrieta et al. (2020); Amodei et al. (2016). Calibration methods further address whether predicted confidence values accurately reflect empirical correctness, with techniques such as temperature scaling improving the correspondence between confidence and accuracy Guo et al. (2017); Ovadia et al. (2019).

Nevertheless, existing approaches primarily address whether predictions are reliable given an assumed representation space. Continual learning asks whether knowledge can be preserved; uncertainty estimation asks how confident a prediction is; trustworthy AI asks whether system behavior is acceptable. None of these explicitly addresses a more fundamental question:

> *Are the distinctions encoded by the current knowledge representation still valid under the present observation conditions?*

This limitation indicates the need for mechanisms that explicitly model knowledge validity under changing observation conditions.

## 2.5 Positioning Epistemic Memory among Existing AI Memory Paradigms

Before formalizing epistemic memory, it is essential to clarify how it relates to and differs from established AI memory paradigms. This positioning prevents misinterpretation and establishes the unique contribution of epistemic memory as an emerging paradigm.

### *Epistemic Memory vs. Continual Learning.*

Continual learning Kirkpatrick et al. (2017); Parisi et al. (2019) primarily addresses the problem of catastrophic forgetting: how to preserve previously learned parameters when acquiring new knowledge. In contrast, epistemic memory addresses a different problem: how to determine whether previously acquired knowledge remains valid under changing observation conditions. A continual learning system may perfectly preserve a learned representation, yet that representation may no longer be semantically justified if the observation process has changed. Continual learning addresses knowledge retention; epistemic memory addresses knowledge validity.

### *Epistemic Memory vs. Uncertainty Estimation.*

Uncertainty estimation methods Gal and Ghahramani (2016); Lakshminarayanan et al. (2017); Ovadia et al. (2019) quantify predictive uncertainty, typically by distinguishing aleatoric uncertainty (inherent noise) from epistemic uncertainty (limited data). However, uncertainty estimation operates within a fixed representation space: it asks how uncertain the model is about predictions given its current representation. Epistemic memory asks a more foundational question: whether the representation itself remains valid. This is reflected in the distinction between uncertainty and epistemic-boundary uncertainty introduced in Section 3.

### *Epistemic Memory vs. World Models.*

World models Ha and Schmidhuber (2018); Hafner et al. (2019) learn latent dynamics for prediction and planning. They excel at forecasting future states from a learned latent representation. However, world models typically assume a stable latent space. Epistemic memory complements world models by maintaining the conditions under which latent representations remain interpretable, effectively providing a validity monitor for the predictive representations that world models provide.

**Table 1** Comparison of AI memory paradigms from the perspective of knowledge validity maintenance. Existing paradigms primarily focus on storing, representing, predicting, or retrieving knowledge, whereas epistemic memory explicitly maintains the validity conditions under which knowledge remains applicable.

| **Paradigm** | **Stored Object** | **Adaptation Mechanism** | **Validity Awareness** | **Representative Examples** |
|---|---|---|---|---|
| Reactive Memory | Sensory states and recent observations | Reactive mapping and feedback control | Implicit | Behavior-based agents Brooks (1991) |
| Representation Memory | Latent features and embeddings | Gradient-based representation learning | Semantic stability assumed | Representation learning and deep neural networks Bengio et al. (2013) |
| Foundation Model Memory | Parametric knowledge in model weights | Pretraining, fine-tuning, and retrieval | Validity implicitly assessed through context | Pretrained language models and foundation models Devlin et al. (2019); Bommasani et al. (2021) |
| Agent Memory | Episodes and retrieved interaction histories | Retrieval, planning, and memory management | Retrieval-oriented | Memory networks, generative agents, and MemGPT Weston et al. (2015); Santoro et al. (2016); Park et al. (2023); Packer et al. (2023) |
| **Epistemic Memory** | **Knowledge, epistemic boundary, and provenance** | **Epistemic boundary tracking and validity maintenance** | **Explicit dynamic validity** | **Observable Belief Memory (OBM), this work** |

### *Epistemic Memory vs. Agent Memory.*

Agent memory architectures Graves et al. (2014); Weston et al. (2015); Packer et al. (2023) store and retrieve information for downstream tasks. They focus on the effective organization and retrieval of experiences. Epistemic memory extends this by maintaining not just the content, but also the conditions under which retrieved knowledge is currently justified, enabling decisions about what to keep, revise, or retire.

**Table 2** Three levels of uncertainty in intelligent systems. Epistemic memory addresses epistemic boundary uncertainty, a distinct level that operates above state and predictive uncertainty.

| Level | Question | Object | Representative Approaches |
|---|---|---|---|
| State uncertainty | What is the hidden state? | World state | Bayesian filtering, POMDPs |
| Predictive uncertainty | How reliable is this prediction? | Model output | Bayesian NNs, ensembles, calibration |
| Epistemic boundary uncertainty | Is the representation still justified? | Semantic distinctions | Epistemic memory, OBM |

## 2.6 Relationship to Uncertainty, Calibration, and Trustworthy AI

To further clarify the distinction between epistemic memory and existing reliability-oriented approaches, it is useful to distinguish three levels of uncertainty that arise in intelligent systems.

### *Level 1: State uncertainty.*

State uncertainty concerns the question: *What is the hidden state of the environment?* This is the classical uncertainty addressed by Bayesian filtering, Kalman filters, and partially observable Markov decision processes Aström (1965); Kaelbling et al. (1998). The object of uncertainty is the world state, and the representation space is assumed fixed.

### *Level 2: Predictive uncertainty.*

Predictive uncertainty concerns the question: *How reliable is this prediction?* This includes aleatoric uncertainty (inherent noise in observations) and epistemic uncertainty (limited data), as addressed by Bayesian neural networks, deep ensembles, and calibration methods Gal and Ghahramani (2016); Kendall and Gal (2017); Lakshminarayanan et al. (2017); Guo et al. (2017); Ovadia et al. (2019). The object of uncertainty is the model output, and the representation space is still assumed fixed.

### *Level 3: Epistemic boundary uncertainty.*

Epistemic boundary uncertainty concerns a different question: *Is the representation itself still justified under current observation conditions?* This is the level addressed by epistemic memory. The object of uncertainty is the validity of the semantic distinctions encoded by the representation. Unlike state and predictive uncertainty, which operate within a fixed representation, epistemic boundary uncertainty addresses whether the representation itself remains meaningful.

Therefore, epistemic memory should not be viewed as another uncertainty estimation technique, but as a mechanism for maintaining the validity boundary within which uncertainty estimates themselves remain meaningful. This distinction is crucial for understanding why epistemic memory addresses a different problem from uncertainty quantification.

## 2.7 Relationship to Metacognition and Cognitive Architectures

The concept of epistemic memory is closely related to metacognition, a long-standing research topic in cognitive science concerned with monitoring and regulating one's own cognitive processes. Human cognition is not limited to representing external information; intelligent behavior also requires awareness of the status, reliability, and limitations of one's own knowledge Flavell (1979); Nelson and Narens (1990).

Cognitive architectures have incorporated related ideas by modeling mechanisms for memory, reasoning, goal management, and self-monitoring. For example, architectures such as ACT-R and SOAR aim to provide unified computational accounts of human cognition by integrating declarative memory, procedural knowledge, and reasoning processes Anderson et al. (2004); Newell (1990). These approaches highlight the importance of internal cognitive states for intelligent behavior and provide important foundations for modeling adaptive agents.

However, existing AI implementations inspired by metacognition generally focus on evaluating the reliability of decisions or predictions rather than explicitly maintaining the validity conditions of the underlying representations. For example, uncertainty estimation methods quantify confidence or predictive uncertainty Gal and Ghahramani (2016); Kendall and Gal (2017), while confidence-aware systems and introspective models assess whether their outputs are likely to be correct Dearden et al. (1999); Cox (2005). These approaches are valuable for reliability assessment, but they typically assume that the representation space in which uncertainty is evaluated remains valid.

Epistemic memory addresses a complementary but distinct problem. Rather than asking only whether an agent is confident about a prediction, epistemic memory asks whether the distinctions encoded by the current representation remain justified under the present observation conditions. In this sense, metacognition concerns awareness and regulation of knowledge states, whereas epistemic memory concerns the maintenance of the validity boundary that determines when those knowledge states remain applicable.

This distinction becomes increasingly important for adaptive and embodied intelligent systems. An agent may preserve accurate memories, maintain high confidence predictions, and perform effective inference under the conditions where its knowledge was acquired. However, changes in sensing configuration, environment, or interaction history may alter which distinctions remain observable. Without an explicit representation of these evolving validity conditions, an agent may continue to apply previously learned knowledge beyond the boundary where it remains justified.

From this perspective, epistemic memory can be viewed as a formal extension of metacognitive capabilities toward dynamic knowledge validity maintenance. It does not replace metacognition or cognitive architectures; rather, it provides an additional structural mechanism for tracking the relationship among knowledge, observation conditions, and representational boundaries. Such a capability is essential for future self-maintaining intelligent systems that must continuously adapt not only their beliefs, but also their understanding of when those beliefs remain valid.

## 3 Formal Foundations of Epistemic Memory

Modern intelligent agents typically represent uncertainty as uncertainty over hidden world states. Classical formulations such as Bayesian filtering and partially observable decision processes describe uncertainty through beliefs over latent states Aström (1965); Smallwood and Sondik (1973); Kaelbling et al. (1998). Embodied agents, however, introduce another fundamental form of uncertainty: epistemic boundary uncertainty—uncertainty over which distinctions between possible states are currently justified by available evidence.

A key distinction is that conventional belief-state formulations assume a fixed state space and observation model, whereas epistemic memory addresses uncertainty over the validity of the representation itself. The object of uncertainty is therefore not only the hidden world state, but the evolving partition through which the agent distinguishes possible states.

This distinction is essential because embodied agents actively modify their observation process through movement, sensing actions, and environmental interaction Bajcsy (1988); Aloimonos et al. (1988); Brooks (1991). The representation structure therefore cannot always be assumed to remain fixed, but should evolve with the changing boundary of observability. We formalize this principle through two coupled equivalence structures: the *observation quotient*, which characterizes distinguishability under a current sensing condition, and the *epistemic quotient*, which characterizes distinguishability given accumulated interaction history and currently available sensing capabilities.

### 3.1 Observation Quotient

Let $\mathcal{H}$ denote the space of hypotheses maintained by the agent and $\mathcal{U}$ the sensing configuration space. A sensing configuration $u \in \mathcal{U}$ induces an observation operator

$$H_u : \mathcal{H} \to \mathcal{Y},$$

which defines the observation distribution of each hypothesis.

**Definition 1** (Observation Equivalence). *Two hypotheses $h_1, h_2 \in \mathcal{H}$ are observationally equivalent under $u$ if*

$$h_1 \sim_u^{obs} h_2 \iff p(y \mid h_1, u) = p(y \mid h_2, u), \qquad \forall y.$$

The corresponding observation quotient is

$$\mathcal{O}_u = \mathcal{H} / \sim_u^{obs}.$$

Each equivalence class represents hypotheses that cannot be distinguished under the current sensing condition. This formulation is related to observation equivalence in partially observable systems Kaelbling et al. (1998). Unlike conventional formulations with fixed observation structures, we consider embodied settings in which the equivalence relation itself evolves as sensing conditions change.

## 3.2 History-Dependent Epistemic Quotient

In embodied interaction, observations accumulate over time. Let

$$\eta_t = (o_{1:t}, a_{1:t-1})$$

denote the interaction history up to time $t$.

Unlike a fixed observation quotient, the epistemic quotient characterizes whether two hypotheses remain distinguishable given acquired evidence and the sensing actions currently available to the agent.

**Definition 2** (Epistemic Quotient)**.** *Let $\mathcal{A}_t^{\text{avail}}$ denote the set of sensing actions currently available to the agent. Given history $\eta_t$, two hypotheses $h_1, h_2 \in \mathcal{H}$ are epistemically equivalent if*

$$h_1 \sim_{\eta_t}^{epi} h_2$$

*when they induce identical next-step observation distributions for every available sensing action:*

$$p(o_{t+1} \mid \eta_t, h_1, a) = p(o_{t+1} \mid \eta_t, h_2, a), \qquad \forall a \in \mathcal{A}_t^{\text{avail}}.$$

The corresponding epistemic quotient is

$$\mathcal{Q}_{\eta_t} = \mathcal{H} / \sim_{\eta_t}^{epi} .$$

Each equivalence class in $\mathcal{Q}_{\eta_t}$ represents hypotheses that cannot currently be distinguished given the agent's accumulated experience and available sensing capability.

The epistemic quotient differs from classical abstraction frameworks. Predictive state representations characterize states through future observable consequences Littman et al. (2001); Singh et al. (2004). Bisimulation-based abstractions preserve behavioral equivalence through transition dynamics and reward structures Givan et al. (2003); Li et al. (2006). Object-centric representations identify entities and semantic components in observations Greff et al. (2019); Locatello et al. (2020). In contrast, the epistemic quotient does not assume a fixed equivalence relation; it models how that relation evolves with sensing capability, interaction history, and available actions.

**Remark 3.** *The epistemic quotient is defined over currently available sensing actions rather than all possible future action sequences. This allows the quotient to evolve when sensing capability changes due to movement, occlusion, or sensor reconfiguration. Defining equivalence over all possible actions would instead reduce the representation to an environment-dependent predictive abstraction rather than an agent-dependent epistemic structure.*

## 3.3 Quotient Dynamics and Refinement

A central property of the epistemic quotient is that it is dynamic. Unlike classical state abstractions, whose equivalence relations are usually defined with respect to fixed transition dynamics or reward structures Givan et al. (2003); Li et al. (2006), the epistemic quotient evolves as the agent acquires evidence or changes its available sensing capabilities.

Additional observations may separate previously indistinguishable hypotheses, whereas degraded sensing may remove previously available distinctions. Quotient evolution therefore represents changing distinguishability rather than a fixed abstraction of the environment.

**Definition 4** (Quotient Refinement)**.** *Given two histories $\eta_t$ and $\eta_{t'}$, where $\eta_{t'}$ contains additional observations compared with $\eta_t$, we say that $Q_{\eta_{t'}}$ is a refinement of $Q_{\eta_t}$ if every equivalence class in $Q_{\eta_t}$ can be represented as a union of equivalence classes in $Q_{\eta_{t'}}$.*

This definition is related to classical notions of abstraction refinement, where additional information produces a finer partition of the underlying hypothesis space Givan et al. (2003); Li et al. (2006). Here, however, refinement is induced by changes in the agent's observation capability rather than changes in the underlying environment model.

For example, suppose two objects are initially indistinguishable because only their silhouettes are observed:

$$\{h_1, h_2\} \in Q_{\eta_t}.$$

After the agent moves to a better viewpoint, their material becomes visible:

$$\{h_1\}, \{h_2\} \in Q_{\eta_{t'}}.$$

The physical states remain unchanged; only the epistemic boundary changes. This provides a theoretical foundation for active perception, where actions are selected not only to acquire information but also to modify future observability Bajcsy (1988); Aloimonos et al. (1988); Lindley (1956); MacKay (2003).

**Remark 5.** *The dynamic nature of $A_t^{\text{avail}}$ enables quotient dynamics. When the agent moves or changes sensing configuration, available observation mechanisms may change, refining, coarsening, or transforming the epistemic quotient. This captures the embodied principle that what an agent can know depends jointly on the environment and its sensing capabilities Brooks (1991); Pfeifer and Bongard (2007).*

## 3.4 Relationship to Existing Frameworks

The epistemic quotient is related to existing representations of uncertainty, prediction, and abstraction, but models a different object. Existing frameworks typically characterize uncertainty or equivalence within a predefined representation space, whereas the epistemic quotient characterizes the evolving boundary that determines which distinctions remain observable to an embodied agent.

### *Belief states in POMDPs.*

A belief state represents uncertainty over hidden physical states:

$$b_t(s) = P(s_t = s \mid \eta_t),$$

where $\eta_t$ denotes the observation-action history. POMDPs provide a principled framework for sequential decision making under partial observability Aström (1965); Smallwood and Sondik (1973); Kaelbling et al. (1998). However, belief states assume a predefined state space and observation model. Their uncertainty concerns *which state is likely*, whereas the epistemic quotient concerns *which distinctions between hypotheses are currently justified*. Thus, belief states represent uncertainty within a fixed semantic space, while the epistemic quotient represents the currently valid structure of distinctions.

### *Predictive state representations.*

Predictive state representations (PSRs) characterize states through conditional predictions of future observations given histories and action sequences Littman et al. (2001); Singh et al. (2004). The epistemic quotient shares the principle that observability should define representation, but the two differ in their target of equivalence. PSRs define equivalence through identical predictive consequences under a specified action-observation protocol:

$$P(o_{\text{future}} \mid \eta, a_{\text{future}}),$$

whereas epistemic equivalence is defined through current distinguishability under available sensing actions:

$$p(o_{t+1} \mid \eta_t, h, a), \quad a \in A_t^{\text{avail}}.$$

Consequently, PSRs model what observations can be predicted, whereas epistemic quotients model which distinctions can currently be established. The latter may change when an embodied agent moves, changes sensors, or encounters new environmental constraints.

### *Bisimulation and state abstraction.*

Bisimulation-based abstractions identify states that are equivalent with respect to transition dynamics and reward structures Givan et al. (2003); Li et al. (2006). These approaches reduce planning complexity by preserving behavioral properties.

However, bisimulation is an environment-centered notion of equivalence:

Do these states behave equivalently?

In contrast, the epistemic quotient is agent-centered and observation-dependent:

Can these hypotheses currently be distinguished by this agent?

Two states may be behaviorally equivalent while remaining distinguishable under a particular sensing capability. Conversely, hypotheses may be epistemically equivalent under limited sensing even when their underlying dynamics differ.

***Summary.***

The relationship among these representations can be summarized as:

| Representation | Fundamental Question |
|---|---|
| Belief state | Which physical state is probable? |
| Predictive state | Which future observations are predictable? |
| Bisimulation | Which states are behaviorally equivalent? |
| Epistemic quotient | Which distinctions are currently justified? |

Thus, the epistemic quotient is not another abstraction of the world state. It is an abstraction of the agent's current boundary of observability, enabling epistemic memory to maintain beliefs as the semantic structure itself evolves during embodied interaction. Unlike belief-state uncertainty, which quantifies uncertainty over states within a predefined representation space, epistemic memory models the evolution of the representation space itself.

# 4 Semantic Coordinate Drift

The epistemic quotient defines the distinctions that an embodied agent can currently justify from available evidence. Most modern representation learning systems, however, assume that semantic coordinates remain stable after learning. A latent dimension, cluster, or embedding direction is expected to preserve consistent meaning across observations and environments Bengio et al. (2013). Here, a semantic coordinate system refers to the equivalence structure induced by the representation rather than the numerical basis of a latent vector space: hypotheses mapped to the same coordinate are treated as semantically equivalent.

This assumption becomes problematic when the epistemic quotient changes. Because embodied agents modify sensing capability through movement, interaction, and sensor configuration, the valid equivalence relation over hypotheses may evolve. A semantic coordinate system that is exact under one epistemic boundary may therefore become inconsistent under another.

We refer to this phenomenon as *semantic coordinate drift*. It does not arise from insufficient model capacity or poor optimization, but from a structural mismatch between a fixed semantic partition and a dynamically evolving epistemic partition.

## 4.1 Quotient-Exact Semantic Representations

Let a semantic representation be

$$R : H \to C,$$

where $\mathcal{H}$ denotes the hypothesis space and $\mathcal{C}$ the semantic coordinate space. Under the quotient-exact criterion, a deterministic representation induces a partition through equality of semantic coordinates.

**Definition 6** (Quotient-Exact Representation)**.** *A semantic representation* $R : \mathcal{H} \to \mathcal{C}$ *is quotient-exact with respect to an epistemic quotient* $\mathcal{Q}_\eta$ *if*

$$R(h_i) = R(h_j) \iff h_i \sim_\eta^{epi} h_j .$$

Quotient exactness requires that the representation preserve exactly the distinctions supported by the current epistemic boundary: indistinguishable hypotheses should not receive different semantic coordinates, while distinguishable hypotheses should not be collapsed.

This differs from conventional representation objectives based on reconstruction, prediction, invariance, or downstream task accuracy Bengio et al. (2013). Such objectives typically assume a stable semantic structure, whereas quotient exactness requires alignment with a changing observation boundary.

## 4.2 Incompatibility of Fixed Semantic Coordinates

We now show that a history-independent semantic coordinate system cannot remain universally valid when epistemic quotients change.

**Proposition 7** (Semantic Coordinate Incompatibility)**.** *Let* $\mathcal{Q}_{\eta_1}$ *and* $\mathcal{Q}_{\eta_2}$ *be epistemic quotients induced by different observation histories. If*

$$\mathcal{Q}_{\eta_1} \neq \mathcal{Q}_{\eta_2} ,$$

*then no single history-independent representation*

$$R : \mathcal{H} \to \mathcal{C}$$

*can be quotient-exact with respect to both.*

*Proof* Assume such a representation $R$ exists. Quotient exactness under $\mathcal{Q}_{\eta_1}$ and $\mathcal{Q}_{\eta_2}$ requires

$$R(h_i) = R(h_j) \iff h_i \sim_{\eta_1}^{epi} h_j$$

and

$$R(h_i) = R(h_j) \iff h_i \sim_{\eta_2}^{epi} h_j .$$

Hence

$$h_i \sim_{\eta_1}^{epi} h_j \iff h_i \sim_{\eta_2}^{epi} h_j ,$$

implying

$$\mathcal{Q}_{\eta_1} = \mathcal{Q}_{\eta_2} ,$$

which contradicts the assumption. □

## 4.3 Quotient Incompatibility Lower Bound

The proposition establishes that exact quotient preservation is impossible when epistemic quotients differ. We next quantify the unavoidable error of any fixed representation.

For any partition $Q$, define the co-membership indicator

$$\chi_Q(h_i, h_j) = \mathbf{1}[h_i \sim_Q h_j].$$

The pairwise disagreement distance between two quotients is

$$d_{\text{pair}}(\mathcal{Q}_1, \mathcal{Q}_2) = \frac{1}{|\mathcal{P}|} \sum_{(i,j)\in\mathcal{P}} |\chi_{\mathcal{Q}_1}(h_i, h_j) - \chi_{\mathcal{Q}_2}(h_i, h_j)|,$$

where $P$ is the set of all unordered hypothesis pairs. This distance is zero for identical quotients and increases with their disagreement.

Let $Q_R$ denote the partition induced by a fixed semantic representation $R$. Define

$$\mathrm{Err}(R;Q) = d_{\text{pair}}(Q_R, Q).$$

**Theorem 8** (Quotient Incompatibility Lower Bound)**.** *Let $Q_i$ and $Q_j$ be two epistemic quotients and $Q_R$ the partition induced by a shared history-independent representation R. Then*

$$\mathrm{Err}(R;Q_i) + \mathrm{Err}(R;Q_j) \geq d_{\text{pair}}(Q_i, Q_j).$$

*Proof* By the triangle inequality of normalized pairwise disagreement,

$$d_{\text{pair}}(Q_i, Q_j) \leq d_{\text{pair}}(Q_i, Q_R) + d_{\text{pair}}(Q_R, Q_j).$$

Substituting the definition of representation error yields the result. □

The theorem strengthens the incompatibility proposition by quantifying the unavoidable inconsistency of a shared semantic representation. As disagreement between epistemic quotients increases, the total representation error across the two regimes must also increase. Thus, increasing model capacity alone cannot eliminate semantic coordinate drift when the representation is constrained to induce a shared history-independent partition.

## 4.4 Degeneracy of Class-Level Transport under Symmetric Crossing Quotients

We next identify a failure mode of quotient-level memory under non-nested epistemic transitions.

**Proposition 9** (Degeneracy of Naive Transport)**.** *Let $Q_i$ and $Q_j$ be two epistemic quotients. Define the naive overlap-based transport kernel*

$$K_{naive}(C_j|C_i) = \frac{|C_i \cap C_j|}{|C_i|}.$$

*If the quotients are* symmetrically crossing *in the sense that*

$$\frac{|C_i \cap C_j|}{|C_i|} = \pi_j, \quad \forall C_i \in \mathcal{Q}_i,$$

*where $\pi_j$ depends only on $C_j$, then*

$$b_j^-(C_j) = \sum_{C_i \in \mathcal{Q}_i} b_i(C_i)\pi_j = \pi_j,$$

*which is independent of the original belief $b_i$.*

*In contrast, provenance-aware transport*

$$K_\rho(C_j|C_i) = \sum_{\xi \in C_i \cap C_j} \rho_i(\xi|C_i)$$

*is not forced to collapse under class-overlap symmetry alone. Because $K_\rho$ depends on the within-class distribution $\rho_i$, the transported belief can remain dependent on pre-transition evidence whenever provenance is non-uniform across crossing intersections.*

*Proof* For naive transport,

$$b_j^-(C_j) = \sum_{C_i} K_{\text{naive}}(C_j|C_i)b_i(C_i) = \sum_{C_i} \pi_j b_i(C_i) = \pi_j$$

because $\Sigma_{C_i} b_i(C_i) = 1$. □

The proposition identifies a class of crossing transitions in which class-level belief alone is insufficient for informative transport. When every old quotient class overlaps new classes in similar proportions, the transported belief becomes independent of the original class-level belief. Within-class provenance $\rho_i$ preserves information that would otherwise be lost when quotient boundaries are reorganized.

## 4.5 Adaptive Representations versus Explicit Epistemic Tracking

The preceding results do not imply that context-conditioned or recurrent representations cannot encode epistemic information. A sufficiently expressive representation

$$z_t = f(\eta_t)$$

may implicitly encode both the world state and properties of the current observation regime.

The distinction is between *adaptive feature representation* and *explicit epistemic structure*. Changes in $z_t$ alone do not reveal which hypothesis distinctions have emerged, disappeared, or been reorganized, nor do they explicitly specify correspondence between semantic identities across changing distinguishability conditions.

For quotient-aware reasoning, the representation should support recovery of the current equivalence structure, or an operationally equivalent object from which quotient transitions can be identified. Epistemic memory therefore does not claim that latent representations are intrinsically insufficient; rather, it makes evolving distinguishability an explicit computational object for quotient estimation, transition detection, and belief transport.

# 5 From Theory to Constructive Realization: Observable Belief Memory

The theoretical framework established above provides a formal foundation for epistemic memory. However, a conceptual framework alone does not constitute an operational architecture. This section introduces *Observable Belief Memory (OBM)* as one possible constructive realization of epistemic memory principles.

OBM serves as an explicit epistemic governance layer that tracks the evolving distinguishability structure of the agent's hypothesis space. OBM is not intended to replace neural representations, world models, or retrieval mechanisms. Instead, it operates as a governance layer that tracks when these representations remain justified under current observation conditions. OBM complements them by maintaining the epistemic structure under which latent information can be consistently interpreted across changing observation conditions.

The central memory state is

$$M_t = (\mathcal{Q}_t, b_t, \rho_t),$$

where:

- $\mathcal{Q}_t$ is the current epistemic quotient;
- $b_t \in \Delta(\mathcal{Q}_t)$ is the belief distribution over quotient classes;
- $\rho_t(\xi / C) = P(\xi / C, \eta_t)$ is the provenance distribution within each class.

While $b_t$ expresses which epistemic class is plausible, $\rho_t$ preserves unresolved hypothesis distributions within each class. This provenance is *not* interpreted as an epistemically justified distinction; it is retained only as a transport scaffold when quotient boundaries later reorganize.

Thus, OBM maintains both uncertainty over candidate hypotheses,

$$P(\xi / \eta_t),$$

and the epistemic structure with respect to which this uncertainty is interpreted:

$$M_t = (Q_t, b_t, \rho_t).$$

This design addresses semantic coordinate drift by allowing the memory representation to evolve with sensing capability.

## 5.1 Dynamic Epistemic Memory Representation

In partially observable environments, the true world state is unavailable. Similar to Bayesian state estimation in POMDPs Aström (1965); Kaelbling et al. (1998), OBM maintains at time $t$ a finite candidate hypothesis set

$$\mathcal{H}_t = \{\xi_t^{(1)}, \ldots, \xi_t^{(N)}\} \subseteq \mathcal{H}.$$

Rather than identifying a single latent state, OBM estimates which hypotheses remain distinguishable under the current observation condition. Given history $\eta_t$, pairwise epistemic distinguishability is

$$d_{ij}^{(t)} = P(\xi_i \not\sim_{\eta_t}^{epi} \xi_j \mid \eta_t).$$

In implementation, this quantity is estimated by

$$d_{ij}^{(t)} = \sigma\left(g_\theta(\phi(\xi_i), \phi(\xi_j), \eta_t, u_t)\right),$$

where $\phi(\cdot)$ is a shared observation encoder and $g_\theta$ predicts whether two hypotheses are distinguishable under the current observation context. The estimator is trained using pairwise distinguishability labels derived from reference quotients or observation-consistency supervision, depending on the experimental setting.

The resulting relation is represented as a weighted hypothesis graph

$$G_t = (\mathcal{H}_t, A_t), \qquad A_{ij}^{(t)} = 1 - d_{ij}^{(t)}.$$

The epistemic quotient is obtained by projecting this estimated relation onto the space of valid equivalence relations. In generic hypothesis spaces, pairwise distinguishability is projected onto a valid partition via spectral clustering; in structured sensing environments, currently observable semantic factors are inferred directly to construct the induced quotient.

This projection enforces a valid equivalence structure. Unlike conventional representation learning methods that optimize feature similarity or task prediction Bengio et al. (2013), OBM explicitly aligns representation structure with current distinguishability.

Because the epistemic quotient evolves over time, beliefs defined on different quotients cannot be directly compared. OBM therefore introduces a belief transport mechanism between changing epistemic spaces.

Given a quotient transition

$$Q_i \to Q_j,$$

belief transport is defined as

$$\Phi_{ij} : \Delta(Q_i) \to \Delta(Q_j),$$

with

$$\Phi_{ij}(b)(C_j) = \sum_{C_i \in Q_i} K(C_j | C_i, e_{i\to j}) b(C_i),$$

where $K$ represents class correspondence across the transition and satisfies

$$\sum_{C_j \in Q_j} K(C_j | C_i, e_{i\to j}) = 1.$$

For refinement transitions,

$$C_i \to \{C_{j1}, C_{j2}, \ldots\},$$

consistency requires

$$K(C_j / C_i) = 0, \quad C_j \not\subseteq C_i.$$

For non-nested transitions, class inclusion is insufficient. The transport kernel is instead constructed through hypothesis-level correspondence:

$$K(C_j | C_i, e_{i\to j}) = \sum_{\xi \in C_i} P(C_j | \xi, e_{i\to j}) P(\xi | C_i).$$

Here, transition evidence $e_{i\to j}$ preserves hypothesis correspondence across reorganized partitions, allowing probability mass to be redistributed without assuming refinement.

Under deterministic correspondence,

$$P(C_j / \xi, e_{i\to j}) = \mathbf{1}[\xi \in C_j],$$

and the general kernel reduces to

$$K_\rho(C_j | C_i) = \sum_{\xi \in C_i \cap C_j} \rho_i(\xi | C_i).$$

The transported prior is

$$b_j^-(C_j) = \sum_{C_i \in Q_i} K_\rho(C_j | C_i) b_i(C_i),$$

which is subsequently updated using the observation likelihood under the new sensing condition.

## 5.2 Posterior-Consistent Update with Provenance

To maintain consistency between class-level belief and within-class provenance, define the hypothesis-level mass

$$p_i(\xi) = b_i(C_i)\rho_i(\xi \mid C_i), \qquad \xi \in C_i.$$

After transport,

$$b_j^-(C_j) = \sum_{\xi \in C_j} p_j^-(\xi),$$

and

$$\rho_j^-(\xi \mid C_j) = \frac{p_j^-(\xi)}{\sum_{\xi' \in C_j} p_j^-(\xi')}$$

for classes with $b_j^-(C_j) > 0$. Zero-mass classes may be assigned an arbitrary normalized reference distribution as they do not affect subsequent inference.

After receiving observation $o_j$,

$$p_j(\xi) \propto p(o_j \mid \xi)p_j^-(\xi),$$

from which $b_j$ and $\rho_j$ are obtained by marginalization and within-class normalization:

$$b_j(C_j) = \sum_{\xi \in C_j} p_j(\xi), \qquad \rho_j(\xi \mid C_j) = \frac{p_j(\xi)}{\sum_{\xi' \in C_j} p_j(\xi')}$$

for $b_j(C_j) > 0$.

This update sequence

$$p_i \rightarrow p_j^- \rightarrow b_j^-, \rho_j^- \rightarrow p_j \rightarrow b_j, \rho_j$$

ensures that OBM maintains posterior consistency between class-level and hypothesis-level beliefs throughout the inference process, rather than treating provenance as auxiliary metadata.

## 5.3 Query-Conditioned Reasoning and Extension to Epistemic Action Selection

OBM memory is independent of individual tasks or queries. A query instead extracts the relevant semantic projection from the current epistemic quotient.

Let query q define a stochastic answer kernel

$$A_{\mathsf{q}}(y \mid C, \eta_t) = P(q(\xi) = y \mid C, \eta_t),$$

which maps each quotient class to a distribution over query answers. The query-conditioned belief is

$$b_t^{\mathsf{q}}(y) = \sum_{C \in \mathcal{Q}_t} A_{\mathsf{q}}(y \mid C, \eta_t) b_t(C).$$

Different tasks can therefore access different semantic views of the same memory without requiring independent representations.

After belief transport, OBM performs Bayesian updating over the current quotient space:

$$b_{t+1}(C') \propto P(o_{t+1}|C', a_{t+1}) \sum_{C} P(C'|C, a_{t+1}) b_t(C).$$

This extends classical Bayesian filtering Kaelbling et al. (1998) from fixed hidden states to dynamic epistemic classes.

When uncertainty remains high, the agent can select sensing actions that refine the relevant epistemic boundary, connecting OBM with active perception Bajcsy (1988). Although active sensing is not evaluated as a closed-loop component in the present experiments, the OBM formulation naturally extends to epistemic action selection through

$$u_t^{\star} = \arg\max_{u} \mathbb{E}_{o_{t+1}} \left[ H(b_t^{q}) - H(b_{t+1}^{q;u}) \right] - \lambda c(u).$$

This objective follows information-gain principles in experimental design and Bayesian active learning Lindley (1956); MacKay (2003). Rather than merely seeking informative observations, OBM selects actions according to how they transform the epistemic quotient and improve the maintenance of valid distinctions.

# 6 Case Studies and Empirical Demonstrations

We provide empirical demonstrations of epistemic memory principles from four complementary perspectives:

1. **Recoverability (Exp1):** Can the epistemic boundary be recovered from observations?
2. **Embodied validity (Exp2):** Does the epistemic boundary change along real embodied sensing trajectories? This experiment is presented as a real-data case study illustrating the phenomenon rather than a quantitative validation of the full framework.
3. **Behavioral consequence (Exp3):** Does explicit epistemic tracking matter when it changes?
4. **Mechanistic necessity (Exp4):** When is provenance required for belief transport?

Together, these demonstrations evaluate progressively stronger claims: recoverability, real-world occurrence, behavioral consequence, and mechanistic necessity.

All experiments compare OBM with representation-based and uncertainty-based baselines under identical observation conditions. Unless otherwise specified, experiments are repeated over five independent random seeds. Exp4 uses ten seeds with 5,000 episodes per $\gamma$ level per seed. Bootstrap confidence intervals are reported for selected effect sizes. Results are reported as mean ± standard deviation with 95% bootstrap confidence intervals where indicated.

## 6.1 EXP1: Demonstration of Dynamic Quotient Properties

### *Objective.*

The first experiment tests whether dynamic epistemic quotients can be recovered directly from observations. The goal is to identify the observation-induced equivalence structure rather than measure downstream task accuracy.

### *Controlled environment.*

We construct a synthetic environment with finite hypothesis space

$$H = \{h_1, \ldots, h_N\}$$

and multiple sensing conditions

$$u \in U,$$

each inducing an observation operator

$$H_u : H \to Y.$$

Different sensing conditions may therefore induce different epistemic quotients,

$$Q_{u_1} \neq Q_{u_2},$$

making the same physical hypotheses distinguishable or indistinguishable depending on the observation condition. The agent receives observations under changing sensing conditions and must infer the corresponding quotient without access to the ground-truth partition.

### *Compared methods.*

We compare four representation strategies:

- **Fixed latent representation:** a single embedding space shared across observation conditions.
- **History-conditioned representation:** conditions on observation history without explicitly modeling equivalence structure.
- **Oracle quotient:** uses the ground-truth epistemic partition as an upper bound.
- **OBM:** learns and tracks the dynamic epistemic quotient.

### *Evaluation metrics.*

Recovered quotient structure is evaluated using Adjusted Rand Index (ARI), Normalized Mutual Information (NMI), and pairwise consistency $F_{\text{pair}}$.

### *Results.*

Figure 4 summarizes the results. OBM consistently achieves higher agreement with the true epistemic quotient than fixed latent representations under changing observation conditions. In contrast, methods that maintain a single semantic coordinate system exhibit increasing inconsistency as the induced equivalence structure changes. These results show that dynamic epistemic boundaries can be empirically recovered from the observation process.

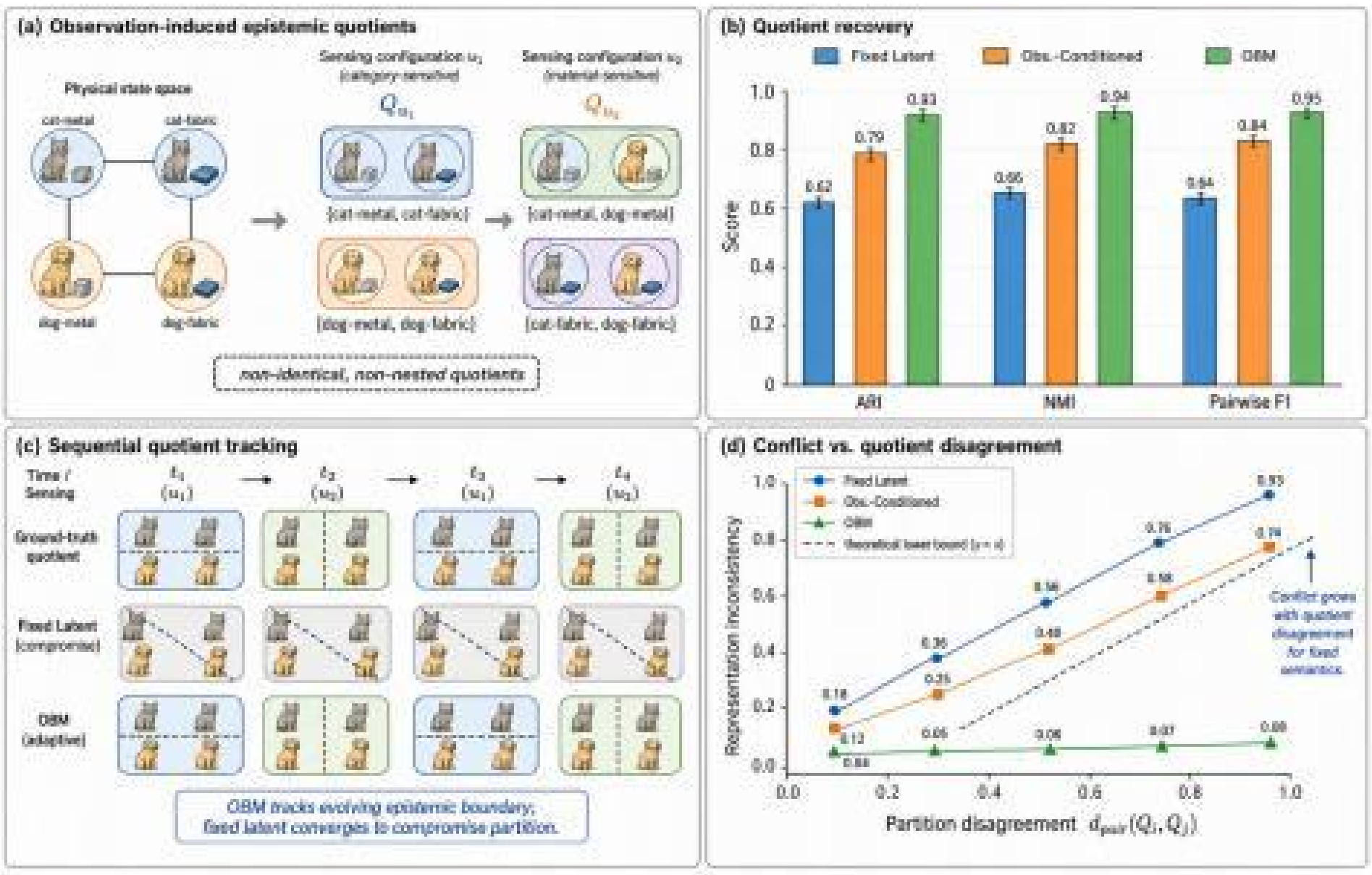

**Fig. 4 Demonstration of dynamic quotient properties and semantic coordinate drift.** (a) Diferent sensing configurations induce diferent equivalence relations over the same hypothesis space. (b) Quotient recovery performance measured by ARI, NMI, and pairwise equivalence F1. OBM more accurately recovers the observation-dependent equivalence structure than fixed latent and history-conditioned representations. (c) Sequential quotient tracking under changing sensing conditions. OBM follows the evolving epistemic boundary, whereas a fixed latent representation converges toward a compromise partition. (d) Representation inconsistency as a function of pairwise partition disagreement $d_{\text{pair}}(Q_i, Q_j)$, consistent with Theorem 8.

## 6.2 EXP2: Embodied Sensing Evidence

#### *Objective.*

This experiment evaluates whether the proposed epistemic memory framework can capture observation-dependent changes in distinguishability under realistic embodied sensing conditions. Unlike controlled simulations where the epistemic quotient is explicitly defined, real sensing environments provide only observations from which an empirical approximation of the observation structure must be inferred.

#### *Dataset and sensing environment.*

The proposed framework is evaluated using the ColoRadar dataset, which provides real-world millimeter-wave radar observations collected in embodied sensing scenarios with synchronized multi-modal sensing information Kramer et al. (2022). The dataset captures environmental variations caused by agent motion, viewpoint changes, and sensor-dependent observation conditions, making it suitable for evaluating whether observable distinctions evolve during interaction with the environment.

### *Experimental setup.*

We evaluate OBM along embodied sensing trajectories extracted from the ColoRadar dataset. Let

$$\tau = (u_1, u_2, \ldots, u_T)$$

denote an observation trajectory, where each $u_t$ represents the sensing condition at time $t$, including viewpoint configuration, sensor geometry, and environmental visibility. At each step, the agent receives an observation

$$y_t \sim p(y|h, u_t),$$

where $h$ denotes the underlying environmental state and $u_t$ determines which aspects of the state remain observable.

### *Empirical observation-induced partition.*

Since the true epistemic quotient is not directly available in real sensing data, we construct an empirical observation-induced partition $\tilde{Q}_t$ from radar observations collected along each trajectory. Specifically, observations are represented using frame-level radar sensing features, and environmental states with similar observation signatures under the current sensing condition are grouped into empirical equivalence classes.

The resulting partition $\tilde{Q}_t$ is treated as an operational approximation of the observation-dependent distinguishability structure rather than the ground-truth epistemic quotient. Therefore, this experiment does not aim to measure exact quotient recovery accuracy. Instead, it evaluates whether real embodied sensing processes induce dynamic changes in observable distinctions and whether epistemic memory can maintain consistent knowledge organization under such transitions.

### *Compared methods.*

We compare fixed representation, history-conditioned representation, and OBM on trajectory-level metrics of partition tracking and belief continuity.

### *Metrics.*

We evaluate three properties.

**Partition consistency:**

$$C_{\text{partition}} = \mathrm{ARI}(\hat{Q}_t, \tilde{Q}_t).$$

**Observation-induced partition transition:**

$$\Delta\tilde{Q}_t = d_{\text{pair}}(\tilde{Q}_t, \tilde{Q}_{t+1}),$$

with predicted transition

$$\overline{\Delta Q_t} = d_{\text{pair}}(\hat{Q}_t, \hat{Q}_{t+1}).$$

We report:

- Mean Absolute Error:

$$E_{\text{transition}} = \frac{1}{T-1}\sum_t \left|\widehat{\Delta Q}_t - \Delta\widetilde{Q}_t\right|;$$

- Spearman rank correlation:

$$\rho_{\text{transition}} = \text{Spearman}(\overline{\Delta Q}_t, \Delta\widetilde{Q}_t).$$

**Belief continuity:**

$$C_{\text{belief}} = 1 - \frac{D_{\text{JS}}(\Phi_{t,t+1}(b_t), \widetilde{b}_{t+1})}{\log 2},$$

where $D_{\text{JS}}$ is the Jensen–Shannon divergence and $\tilde{b}_{t+1}$ denotes an empirical belief derived from the available observations under the current sensing condition. This metric measures whether transported beliefs remain consistent with the observation-supported belief after the epistemic boundary changes.

***Results.***
Figure 5 summarizes the case study results. The same physical environment exhibits different observation-induced partition structures under changing sensing conditions, supporting the view that distinguishability depends on the agent's sensing configuration. The case study illustrates differences in partition tracking and belief continuity across the evaluated representations on this trajectory.

### 6.3 EXP3: Necessity of Explicit Epistemic Memory under Non-Nested Boundary Transitions

***Objective.***

Experiments 1 and 2 establish that epistemic quotients can be recovered from observations and that changes in sensing conditions induce transitions between distinct knowledge partitions. We now investigate whether explicit epistemic memory provides capabilities beyond those of conventional implicit memory mechanisms.

Specifically, this experiment examines whether recurrent latent models and predictive world models can implicitly preserve the information required for knowledge transport across non-nested epistemic transitions, or whether explicit maintenance of epistemic state variables—including quotient structure and within-class provenance—is necessary for reliable belief propagation.

Our central hypothesis is that while implicit memory mechanisms can adapt temporal representations to changing observation conditions, they cannot explicitly preserve the evidence distribution required to reconstruct beliefs after changes in observation-induced partitions. This limitation becomes critical under non-nested boundary transitions, where simple aggregation of class-level beliefs is insufficient and within-class provenance is essential for informative belief transport.

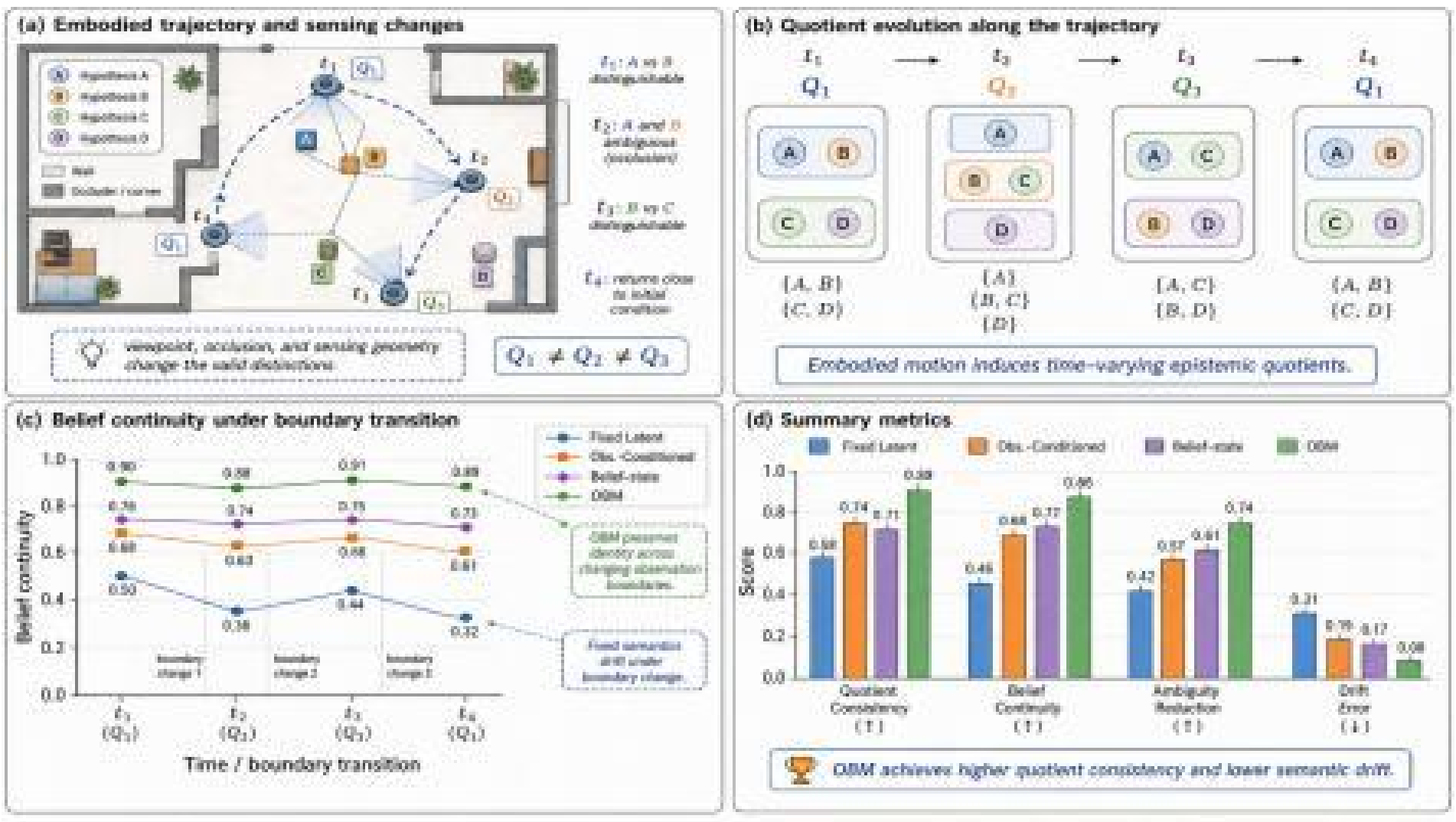

**Fig. 5 Embodied sensing evidence: evolution of observation-induced partitions along real trajectories.** (a) A representative embodied trajectory where viewpoint, sensing geometry, and occlusion induce changing observation conditions and evolving observation-induced partitions. (b) The same physical hypotheses exhibit diferent empirical equivalence structures over time, demonstrating that distinguishability is observation-dependent. (c) The trajectory illustrates how belief continuity difers across the evaluated representations under changing observation conditions. (d) Comparison of partition consistency, transition tracking, and belief continuity across the evaluated representations on this trajectory.

### *Controlled Sequential Sensing Environment.*

We construct a controlled environment with a hidden state

$$h = (c, m, d), \tag{1}$$

where $c$, $m$, and $d$ denote category, material, and distractor attributes. The downstream query is defined as

$$q(h) = (c, m), \tag{2}$$

yielding 16 possible outcomes.

Each sensing condition is parameterized by

$$u = (w_c, w_m, w_d, \theta), \tag{3}$$

where $w_c$, $w_m$, and $w_d$ control the observability of each factor, and $\theta$ represents a sensing-dependent feature transformation. Observations are generated from factor-specific semantic embeddings under a fixed sensing-energy constraint. A learned visibility estimator predicts the currently observable factors,

$$g_\psi(u) \rightarrow (\hat{v}_c, \hat{v}_m, \hat{v}_d), \tag{4}$$

from which the epistemic quotient $\widehat{Q}_t$ is constructed. The estimator is trained independently on 5,000 sensing actions (4,000 training, 500 validation, and 500 held-out), separate from the downstream evaluation trajectories.

Across all five seeds, factor-level and exact visibility-mask prediction achieve 100% accuracy, yielding

$$\mathrm{ARI}(\widehat{Q}_t, Q_t) = 1.0 \tag{5}$$

on held-out transitions. This ensures that differences among the following ablations cannot be attributed to quotient-recovery errors.

### *Non-nested Epistemic Transitions.*

Different sensing conditions reveal different factors, inducing different partitions of the same hypothesis space. For example, category-sensitive sensing induces

$$Q_{\text{cat}} = \{\{\text{cat}-\text{metal}, \text{cat}-\text{fabric}\}, \{\text{dog}-\text{metal}, \text{dog}-\text{fabric}\}\}, \tag{6}$$

whereas material-sensitive sensing induces

$$Q_{\text{mat}} = \{\{\text{cat}-\text{metal}, \text{dog}-\text{metal}\}, \{\text{cat}-\text{fabric}, \text{dog}-\text{fabric}\}\}. \tag{7}$$

These quotients are non-nested,

$$Q_{\text{cat}} \not\preceq Q_{\text{mat}}, \qquad Q_{\text{mat}} \not\preceq Q_{\text{cat}}, \tag{8}$$

meaning belief cannot generally be transferred through simple aggregation or refinement.

We retain sensing-condition pairs whose oracle task performance lies within a controlled range to reduce confounding between quotient incompatibility and intrinsic task difficulty. This yields three attainable disagreement regimes,

$$d_{\text{pair}} \approx 0.194, \quad 0.387, \quad 0.516, \tag{9}$$

denoted low, moderate, and high disagreement. Here $d_{\text{pair}}$ denotes the pairwise disagreement distance between quotients as defined in Section 4. Thirty candidate sensing pairs were generated per incompatibility bin and partitioned into training, validation, and held-out test sets; six held-out pairs per regime were used for final testing over five random seeds. For each held-out sensing pair, performance is evaluated over 1,000 independent test episodes.

### *Compared Methods.*

We compare five memory architectures that span the spectrum from implicit temporal memory to explicit epistemic state tracking, as summarized in Table 3.

**Table 3** Methods evaluated under non-nested quotient transitions: comparison of implicit and explicit memory mechanisms.

| Method | Temporal Memory | Dynamic Quotient | Provenance | Explicit Validity |
|---|---|---|---|---|
| Fixed Latent | No | No | No | No |
| Recurrent Latent | Yes | No | No | No |
| World-Model Latent | Yes | No | No | No |
| OBM-Reset | No | Yes | No | Yes |
| OBM-Naive Transport | Yes | Yes | No | Yes |
| OBM-Provenance Transport | Yes | Yes | Yes | Yes |
| Oracle | Yes | Ground Truth | Ground Truth | Ground Truth |

**Fixed Latent Representation (FLM)** maintains a time-independent embedding $z_t = f_\theta(o_t)$ without any temporal memory, quotient tracking, or provenance preservation. This baseline establishes the lower bound for representation-based approaches.

**Recurrent Latent Model (RLM)** uses a recurrent state $h_t = f_\theta(h_{t-1}, o_t)$ that accumulates temporal information. This baseline represents implicit memory without explicit epistemic structure.

**World-Model Latent (WML)** employs a predictive latent dynamics model $h_t = f_\theta(h_{t-1}, a_{t-1})$ with a learned observation decoder. This baseline represents a representative predictive latent memory architecture. Although sufficiently expressive world models may implicitly encode observation conditions, they do not expose an explicit variable corresponding to epistemic validity.

**OBM-Reset** discards the previous quotient-level belief and initializes the new quotient with an uninformative prior, isolating the contribution of retaining epistemic history rather than the benefit of quotient awareness alone. This baseline recovers the boundary structure immediately (as it explicitly maintains $Q_t$) but cannot preserve belief information across the transition.

**OBM-Naive Transport** preserves the class-level belief $b_i$ but assumes a uniform distribution within each old quotient class: $K_{\text{naive}}(C_j \mid C_i) = |C_i \cap C_j|/|C_i|$. This ablation tests whether class-level belief alone is sufficient for knowledge transport.

**OBM-Provenance Transport** additionally preserves the within-class provenance $\rho_i(h \mid C_i)$: $K_\rho(C_j \mid C_i) = \sum_{h \in C_i \cap C_j} \rho_i(h \mid C_i)$. The transported prior $b_{\bar{j}}(C_j) = \sum_{C_i \in Q_i} K_\rho(C_j \mid C_i) b_i(C_i)$ is subsequently updated using the new observation likelihood.

**Oracle** uses ground-truth quotient information and represents an upper bound on performance.

### *Evaluation Metrics.*

The purpose of this comparison is not to demonstrate universal performance superiority, but to evaluate whether epistemic validity can be explicitly represented and maintained. We therefore evaluate four complementary aspects of knowledge maintenance:

*Validity boundary error* $E_Q = d_Q(\hat{Q}_t, Q_t)$ measures the accuracy with which each method recovers the current epistemic quotient. Here $d_Q$ denotes the normalized pairwise disagreement distance between quotient partitions, defined analogously to the quotient incompatibility metric $d_{\text{pair}}$ in Section 4. This metric directly assesses whether the agent explicitly tracks which distinctions are currently justified.

*Boundary recovery delay* $T_Q = \min_t \{E_Q < \epsilon\}$ measures how quickly each method adapts to the new epistemic boundary following a quotient transition.

*Transport error* $D_{\text{KL}}$ quantifies the divergence between the transported belief and the optimal belief under the new quotient:

$$D_{\text{KL}} = D_{\text{KL}}(b_j^* \parallel b_{j^-}), \tag{10}$$

where $b_j^*$ is the belief that would result from perfect transport given full knowledge of the transition. This metric isolates the information loss incurred during boundary transitions.

*Task success* $S = P(\hat{q} = q(h))$ measures the primary downstream performance.

To quantify the value of explicit epistemic components, we define:

- $\Delta_{\text{retention}} = S_{\text{Prov}} - S_{\text{Reset}}$ (total benefit of epistemic memory),
- $\Delta_{\text{provenance}} = S_{\text{Prov}} - S_{\text{Naive}}$ (benefit of within-class provenance),
- $\Delta_{\text{explicit}} = S_{\text{Prov}} - S_{\text{Recurrent}}$ (benefit of explicit over implicit memory).

### Results.

Figure 7 presents the results across three quotient-incompatibility regimes.

*Validity Boundary Recovery.* Table 4 reports recovery delay $T_Q$ and steady-state boundary error $E_Q$. OBM-based methods (including Reset and Naive Transport) recover the epistemic quotient immediately upon transition ($T_Q = 1$ step), as they explicitly maintain $Q_t$ as a state variable. This demonstrates that explicit quotient tracking provides immediate awareness of boundary changes. However, OBM-Reset recovers the boundary structure without preserving belief information, highlighting the distinction between *boundary awareness* and *belief transport*.

In contrast, Fixed Latent provides no quotient representation, and both Recurrent and World-Model Latent require multiple time steps (2.8–4.1 steps) to implicitly adapt their latent representations to the new observation-induced partition. Steady-state boundary error is $E_Q < 0.01$ for OBM methods versus $E_Q > 0.15$ for implicit baselines, indicating that implicit memory mechanisms never fully recover the exact epistemic structure.

*Transport Error.* OBM-Provenance Transport maintains consistently lower transport error ($D_{\text{KL}} < 0.05$ across all regimes) than implicit baselines ($D_{\text{KL}} > 0.20$), with the gap widening as quotient incompatibility increases. Naive transport provides minimal improvement over reset, as symmetric class overlaps cause $K_{\text{naive}}(C_j \mid C_i) \approx 1/|Q_j|$, making the transported belief nearly independent of the original class-level belief. Provenance-aware transport, by contrast, preserves within-class evidence that enables accurate redistribution under the new quotient.

**Table 4** Boundary recovery delay $T_Q$ and steady-state boundary error $E_Q$ across methods and disagreement regimes.

| Method | $T_Q$ (steps) | | | $E_Q$ (steady) | |
|---|---|---|---|---|---|
| | Low | Moderate | High | Low | High |
| Fixed Latent | > 10 | > 10 | > 10 | 0.48 | 0.52 |
| Recurrent Latent | 3.2 | 3.8 | 4.1 | 0.21 | 0.28 |
| World-Model Latent | 2.8 | 3.5 | 3.9 | 0.19 | 0.25 |
| OBM-Reset | 1.0 | 1.0 | 1.0 | 0.00 | 0.00 |
| OBM-Naive Transport | 1.0 | 1.0 | 1.0 | 0.00 | 0.00 |
| OBM-Provenance Transport | 1.0 | 1.0 | 1.0 | 0.00 | 0.00 |

*Task Success.* Across low, moderate, and high disagreement regimes, Fixed Latent achieves 0.322, 0.267, and 0.158, confirming that a static representation is insufficient for non-nested transitions. Recurrent Latent achieves 0.828, 0.877, and 0.695, and World-Model Latent achieves 0.821, 0.870, and 0.696. OBM-Reset achieves 0.513, 0.428, and 0.229, while OBM-Naive Transport yields 0.550, 0.428, and 0.229. OBM-Provenance Transport achieves 0.842, 0.884, and 0.713, with corresponding provenance gains $\Delta_{\text{provenance}} = 0.292, 0.456, 0.484$ (29.2, 45.6, 48.4 percentage points).

*Implicit vs. Explicit Comparison.* The key insight from these results is not that OBM universally outperforms expressive latent models by a large margin, but rather that OBM achieves competitive performance *while providing explicit interpretability and governance of the epistemic boundary*. Recurrent and world-model baselines achieve competitive average task performance at low disagreement, yet they do not explicitly represent the evolving epistemic boundary that determines when previously learned distinctions remain valid. Their performance degradation under increasing quotient incompatibility emerges implicitly from latent dynamics, whereas OBM maintains the corresponding epistemic state explicitly. This is evidenced by the immediate boundary recovery ($T_Q = 1$) and near-zero boundary error ($E_Q \approx 0$) achieved by OBM, compared with delayed and incomplete recovery by implicit baselines.

The explicit advantage $\Delta_{\text{explicit}} = S_{\text{Prov}} - S_{\text{Recurrent}}$ is 0.014, 0.007, and 0.018—modest in absolute terms but structurally significant. The explicit representation enables immediate boundary adaptation, accurate belief transport, and provenance-aware reasoning that implicit mechanisms cannot provide without additional structural assumptions. The boundary recovery metrics (Table 4) demonstrate this structural advantage directly: OBM knows when the epistemic boundary has changed, whereas implicit methods must infer this change indirectly from downstream prediction errors.

### Discussion.

These results support three conclusions. First, explicit epistemic state tracking provides a distinct computational capability beyond temporal memory accumulation and latent prediction alone. This is evidenced by the immediate boundary recovery ($T_Q$ = 1) and near-zero boundary error ($E_Q \approx 0$) achieved by OBM, compared with delayed and incomplete recovery by implicit baselines.

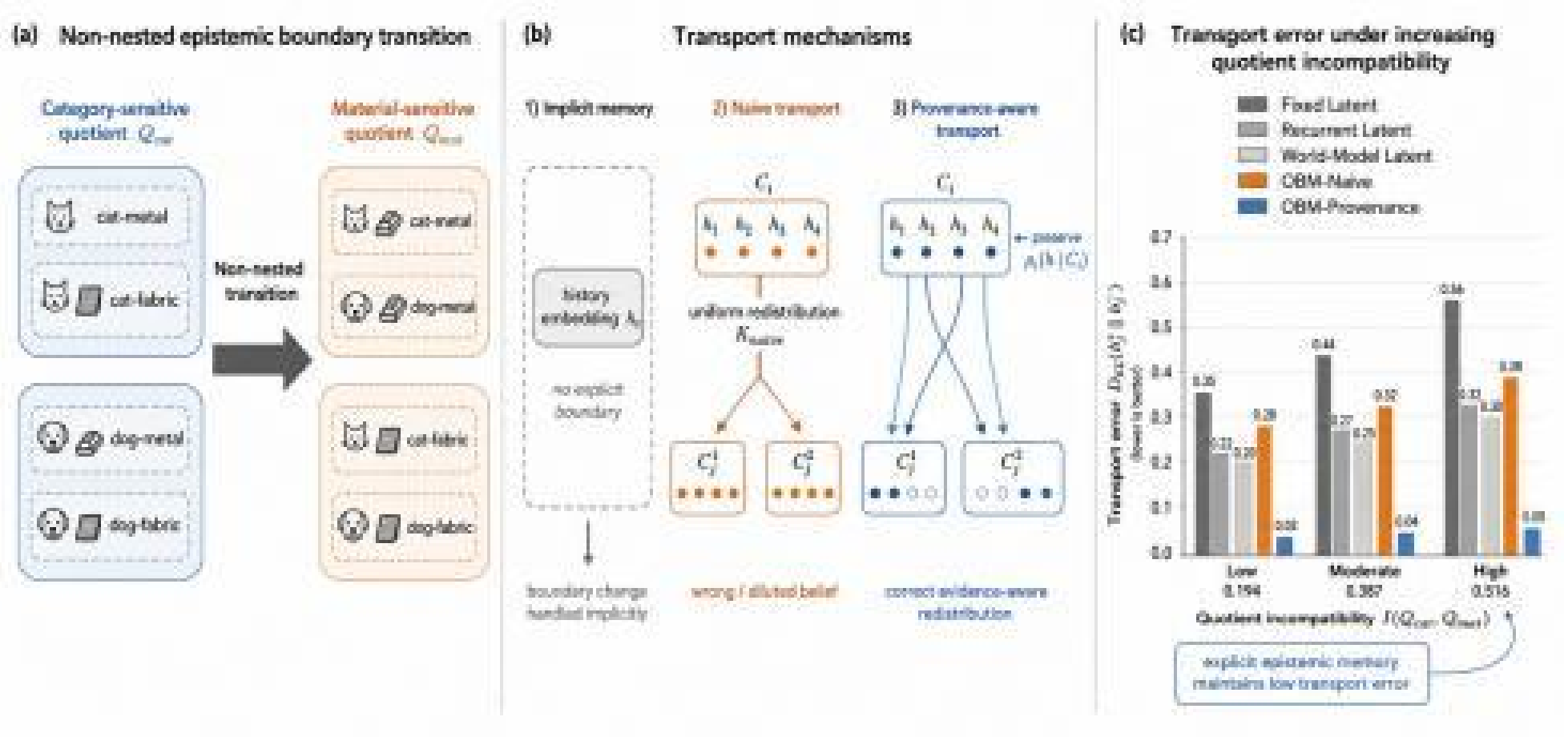

**Fig. 6 Explicit epistemic memory reduces belief transport error under non-nested boundary transitions.** (a) Observation changes induce transitions between incompatible epistemic quotients. (b) Implicit memory mechanisms lack an explicit representation of the changing validity boundary, whereas provenance-aware transport preserves within-class evidence. (c) Quantitative comparison of transport error across increasing quotient incompatibility levels.

Second, within-class provenance is essential for reliable belief transport across non-nested epistemic boundaries, confirming the theoretical prediction of Proposition 9. The provenance gains of 29.2–48.4 percentage points demonstrate that class-level belief alone is insufficient when quotient boundaries are reorganized.

Third, the explicit representation of epistemic boundaries enables immediate adaptation to new observation conditions, whereas implicit mechanisms require multiple time steps to adjust their representations. This suggests that explicit epistemic memory complements rather than replaces implicit memory mechanisms. When the observation-induced semantic structure is stable, implicit latent representations may suffice. However, when sensing conditions change and epistemic boundaries are reorganized, explicit tracking of quotient structure and provenance provides capabilities that cannot be fully replicated by implicit temporal memory alone.

## 6.4 EXP4: Provenance Necessity Stress Test

To test the mechanism predicted by Proposition 9, we systematically vary overlap symmetry between quotients. For the balanced equal-cardinality quotient family considered here, symmetric crossing corresponds to uniform target overlap. We define the normalized overlap-asymmetry score

$$\gamma = \frac{\frac{1}{|\mathcal{Q}_i|}\sum_{C_i \in \mathcal{Q}_i} D_{\mathrm{TV}}\left(K_{\mathrm{naive}}(\cdot \mid C_i), U(\mathcal{Q}_j)\right)}{1 - \frac{1}{|\mathcal{Q}_j|}}. \tag{11}$$

Here $D_{\mathrm{TV}}$ denotes total variation distance, $K_{\mathrm{naive}}(\cdot \mid C_i)$ is the naive transport distribution, and $U(Q_j)$ is uniform over $Q_j$. Unlike $d_{\mathrm{pair}}$, which measures overall partition disagreement, $\gamma$ isolates the symmetry of class-level correspondence. Within this controlled family, $\gamma = 1$ denotes the maximally aligned endpoint and $\gamma = 0$ perfect symmetric crossing.

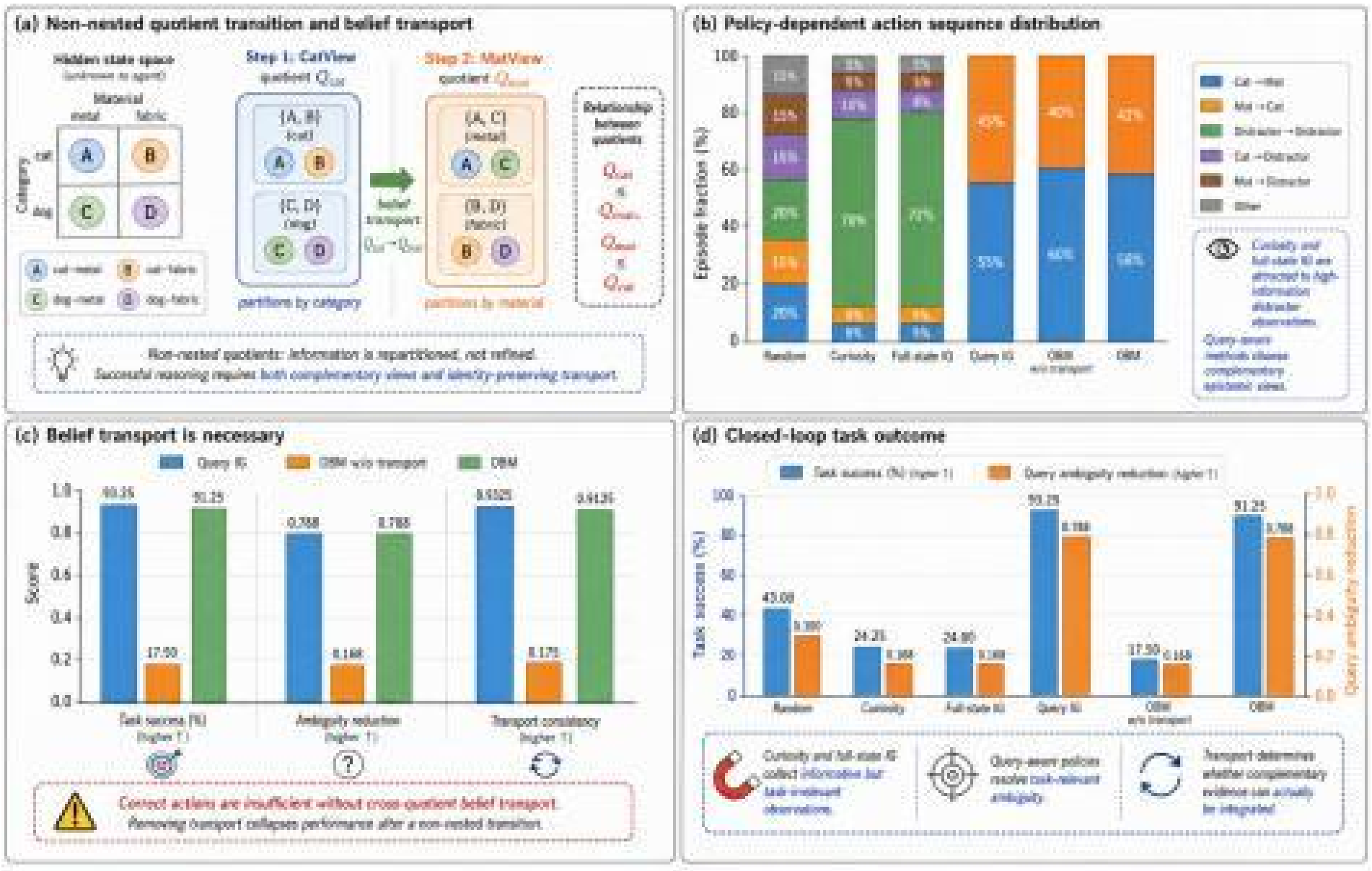

**Fig. 7 Comparison between implicit memory mechanisms and explicit epistemic memory under non-nested quotient transitions.** (a) Example of a non-nested transition from a category-sensitive quotient to a material-sensitive quotient. (b) Validity boundary recovery: OBM-based methods recover the epistemic quotient immediately ($T_Q = 1$), while implicit baselines require 2.8–4.1 steps to adapt their latent representations. (c) Task success comparison across five memory architectures and three quotient-incompatibility regimes. OBM-Provenance Transport maintains competitive performance while providing explicit epistemic governance. (d) Provenance gains of 29.2–48.4 percentage points across the evaluated quotient-incompatibility regimes, demonstrating the importance of within-class provenance for reliable belief transport.

**Table 5** Provenance necessity under controlled overlap symmetry.

| $\gamma$ | $S_{\text{Reset}}$ | $S_{\text{Naive}}$ | $S_{\text{Prov}}$ |
|---|---|---|---|
| 1.00 | 0.553 | 0.948 | 0.948 |
| 0.75 | 0.552 | 0.788 | 0.911 |
| 0.50 | 0.556 | 0.665 | 0.883 |
| 0.25 | 0.551 | 0.581 | 0.864 |
| 0.00 | 0.555 | 0.555 | 0.862 |

At $\gamma = 1$, $S_{\text{Naive}} = S_{\text{Prov}} = 0.948$. At $\gamma = 0$, $S_{\text{Reset}} = S_{\text{Naive}} = 0.555$, whereas $S_{\text{Prov}} = 0.862$, corresponding to a 30.8-percentage-point gain (95% episode-level bootstrap CI: 30.4–31.2 percentage points).

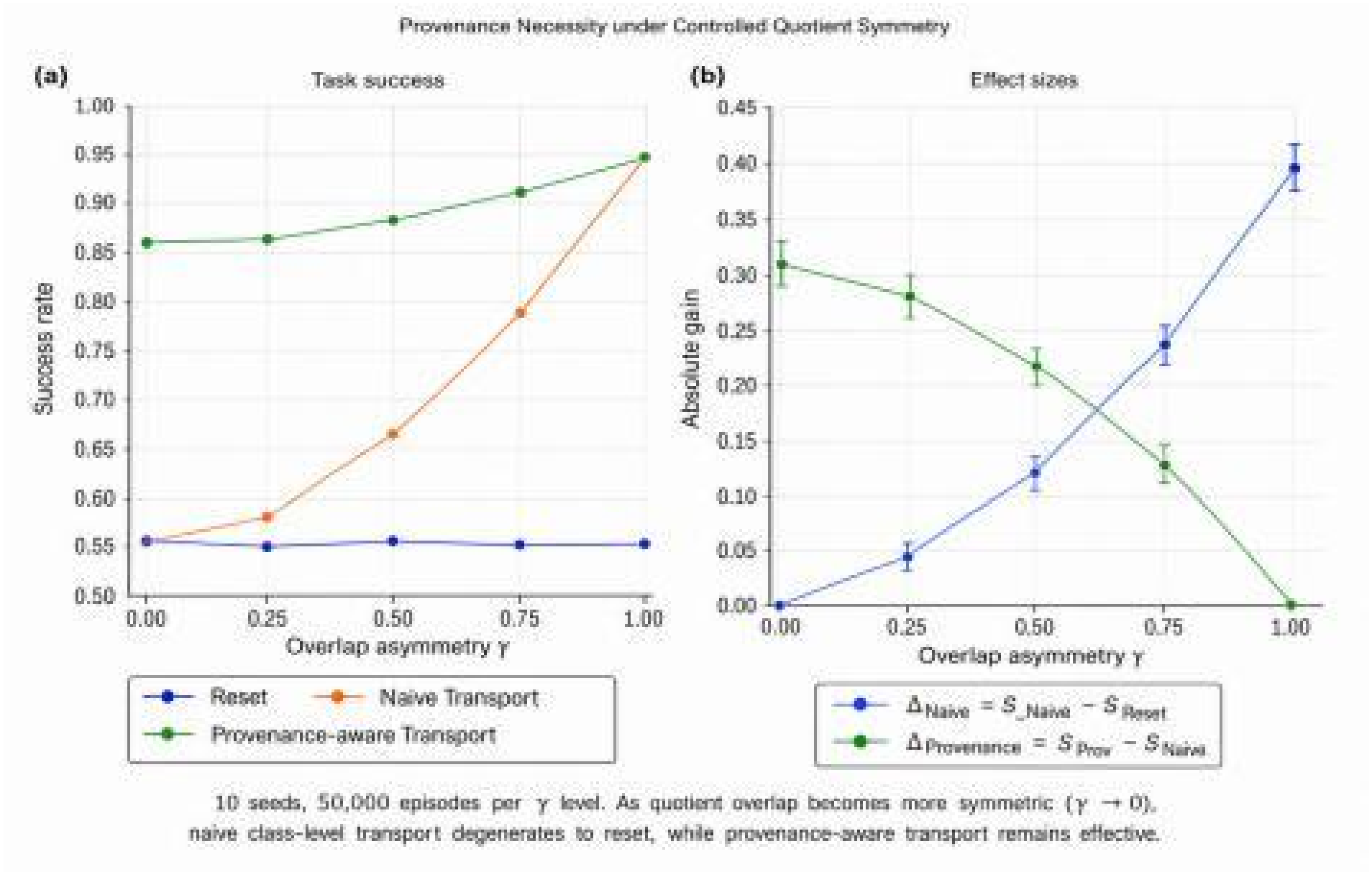

**Fig. 8 Provenance necessity under controlled overlap symmetry.** Within this controlled family, as quotients become more symmetrically crossing ($\gamma \to 0$), naive transport converges to the reset baseline ($S_{\text{Naive}} - S_{\text{Reset}} \to 0$), whereas provenance-aware transport maintains a positive advantage ($S_{\text{Prov}} - S_{\text{Naive}} > 0$). The provenance gain increases with correspondence symmetry.

These results are consistent with Proposition 9: at perfect symmetric crossing, class-level overlap transport becomes independent of the pre-transition belief, whereas provenance-aware transport preserves within-class information lost by this class-level transport mechanism.

# 7 Future Directions: Toward Self-Maintaining AI Architectures

The conceptual framework and constructive demonstrations presented in this work open several promising research directions. Rather than providing a complete solution, this section outlines key areas where epistemic memory principles may be extended and integrated with broader AI research.

## 7.1 Foundation Model Integration

Foundation models represent a major advance in artificial intelligence by providing large-scale parametric knowledge and powerful reasoning capabilities Bommasani et al. (2021). However, the knowledge encoded in foundation models is primarily optimized for prediction and generation. The validity of such knowledge under changing contexts, environments, and observation conditions is generally not explicitly represented.

Epistemic memory offers a complementary direction for future foundation-model-based systems. Foundation models provide knowledge content; epistemic memory provides validity governance. A key research direction is developing interfaces between continuous LLM representations and discrete epistemic quotient structures.

For retrieval-augmented generation (RAG), epistemic memory could extend the current retrieve-generate pipeline to retrieve-validate-organize. Retrieved documents would be assessed against current validity conditions before being used for reasoning or generation. This would enable AI systems to maintain not only what information is available, but whether it remains justified for the current task and context. Current foundation models lack an explicit mechanism for assessing when their parametric knowledge remains valid; epistemic memory provides a governance layer for such assessment.

## 7.2 Embodied Agents and Robotics

Robotics is a natural application domain for epistemic memory. Mobile robots experience changing viewpoints, lighting conditions, sensor configurations, and environmental occlusions. A robot that tracks the epistemic quotient can maintain consistent semantic understanding despite changing observation conditions.

Active perception becomes particularly powerful when integrated with epistemic memory. Rather than selecting actions solely for information gain, the agent can select actions that refine the epistemic boundary, enabling more reliable long-term operation. This direction connects epistemic memory with active sensing, information-theoretic exploration, and adaptive sensing strategies Bajcsy (1988); Aloimonos et al. (1988).

Future work could implement closed-loop epistemic action selection, where agents choose sensing actions based on their expected impact on the epistemic boundary rather than immediate task performance. This would enable robots to actively maintain the validity of their knowledge representations through strategic sensing.

## 7.3 World Models and Predictive Latent Dynamics

World models learn predictive latent dynamics of the environment Ha and Schmidhuber (2018); Hafner et al. (2019, 2020, 2023). They serve as predictive simulators that forecast future states and enable planning. However, world models typically assume that latent distinctions remain stable across time.

A promising direction is integrating epistemic memory with world models to validate latent coordinates. When the agent's sensing condition changes, the world model's latent states may no longer correspond to the same semantic distinctions. Epistemic memory could provide a validity monitor for the world model's latent space, enabling the world model to adapt its predictions when the epistemic boundary shifts.

World Model → Predict latent dynamics
Epistemic Memory → Validate latent coordinates

This integration would enable more robust planning and decision-making under changing observation conditions.

### 7.4 Hardware and Edge Intelligence

The computational demands of epistemic memory—maintaining dynamic quotients, transporting beliefs across changing partitions, and performing posterior-consistent updates—suggest that specialized hardware may be beneficial for scalable deployment. Emerging memory-centric architectures, including processing-in-memory (PIM) and compute-in-memory (CIM), could provide efficient primitives for the graph-structured computations required for quotient inference and belief transport.

Future intelligent hardware may require validity-aware memory primitives that explicitly support the dynamic reorganization of knowledge structures. This suggests a broader shift in memory hierarchy design:

$$\text{Traditional: Data} \rightarrow \text{Memory} \rightarrow \text{Compute}$$
$$\text{Epistemic: Data} \rightarrow \text{Knowledge} \rightarrow \text{Validity} \rightarrow \text{Memory} \rightarrow \text{Compute}$$

This perspective connects epistemic memory with ongoing research in memory-centric computing and neuromorphic architectures, potentially enabling efficient realization of epistemic memory in edge and embedded systems.

### 7.5 Learning Epistemic Boundaries from Experience

The current framework assumes that the observation-induced quotient structure can be estimated from observations, but does not specify a general method for automatically discovering epistemic boundaries directly from high-dimensional sensory streams. Future research should investigate learnable quotient inference mechanisms that can operate in complex and continuously evolving environments.

A promising direction is integrating epistemic boundary discovery with causal representation learning, enabling systems to infer both the world model and its validity boundary simultaneously from experience. This would move beyond the assumption of known factor observability and enable epistemic memory to operate in open-world settings.

## 8 Discussion: Toward Self-Maintaining Intelligent Systems

The results and directions presented in this work suggest that future intelligent systems may benefit from mechanisms that not only store and utilize knowledge, but also maintain the validity conditions under which that knowledge remains applicable. Existing AI systems have achieved remarkable progress in representation learning, foundation models, and autonomous agents; however, these systems generally treat the validity of learned representations as an implicit assumption. Epistemic memory introduces an additional perspective: knowledge should be maintained together with the observation conditions, evidential basis, and representational boundaries that justify its use.

This perspective does not imply replacing existing memory architectures. Instead, epistemic memory provides a complementary governance layer that operates above neural representations, retrieval mechanisms, and world models.

## 8.1 Epistemic Memory as a Knowledge Governance Layer

Modern AI memory mechanisms primarily address three fundamental questions: what information has been stored, how it can be retrieved, and how it can be reused for downstream reasoning. Episodic memories preserve previous experiences, neural representations encode semantic structures, and foundation models store large-scale knowledge within distributed parameters. However, these mechanisms generally assume that the semantic meaning of stored knowledge remains stable across future contexts.

Epistemic memory addresses a complementary question:

Under what conditions does previously acquired knowledge remain justified?

Therefore, epistemic memory should not be interpreted as a replacement for neural memory architectures. Rather, it provides a governance layer that maintains the validity of existing memory representations. In the proposed framework, Observable Belief Memory (OBM) does not replace latent representations, retrieval systems, or predictive models; instead, it maintains the epistemic state associated with those representations, including the current observation-induced quotient, belief distribution, and evidential provenance.

This distinction is important because intelligent systems operating in dynamic environments face a different failure mode from conventional memory systems. A system may successfully retrieve previously learned knowledge while the conditions that originally justified that knowledge have changed. Epistemic memory provides a mechanism for detecting such situations and enabling knowledge reorganization when necessary.

## 8.2 Limitations

While this work establishes the conceptual and formal foundations of epistemic memory, several limitations should be acknowledged.

#### *Controlled environment validation.*

The empirical demonstrations presented in this work are conducted in controlled and structured environments. While the ColoRadar case study provides evidence of boundary evolution along real embodied trajectories, the full framework has not been validated in large-scale, open-world settings with continuous, high-dimensional sensory inputs. Scaling quotient inference to such environments remains an open challenge.

#### *Quotient estimation assumption.*

The current framework assumes that the observation-induced quotient structure can be estimated from observations. While this assumption enables formal analysis, developing scalable methods for automatically discovering epistemic boundaries directly from high-dimensional sensory streams in real-world environments remains an open challenge.

***Open-world scalability.***

The formalization assumes a finite hypothesis space $H$ with structured relations. Real-world intelligent systems operating in open environments may encounter new objects, concepts, and hypotheses that were not present in the initial hypothesis space. The $O(N^2)$ pairwise distinguishability estimation required for quotient inference may become computationally prohibitive for large-scale systems with continuous hypothesis spaces.

# 9 Conclusion

Artificial intelligence has progressed through three memory paradigms: reactive systems that store immediate experiences, representation learning that encodes semantic structures, and foundation models that leverage vast parametric knowledge. This paper has argued that the evolution toward self-maintaining intelligent systems motivates a complementary direction—epistemic memory—that explicitly tracks the validity conditions of stored knowledge.

We formalized the dynamic epistemic quotient, an observation-induced equivalence structure over hypotheses that evolves with an agent's interaction history and sensing capabilities. We derived a pairwise incompatibility lower bound showing that any fixed semantic representation must incur unavoidable error across changing epistemic boundaries. We further identified a failure mode under symmetric crossing quotients, where overlap-based class-level transport becomes independent of pre-transition belief.

We introduced Observable Belief Memory (OBM) as a constructive governance architecture, with state $M_t = (Q_t, b_t, \rho_t)$, combining the current epistemic quotient, belief over quotient classes, and within-class provenance through a posterior-consistent update. Controlled experiments, a real embodied sensing case study on ColoRadar, non-nested transition benchmarks, and a controlled symmetry stress test demonstrated that dynamic distinguishability can be explicitly tracked.

Across three evaluated transition regimes, provenance-aware transport improved over naive transport by 29.2–48.4 percentage points. Under perfect symmetric crossing, the advantage was 30.8 percentage points, while performance remained competitive with recurrent and world-model baselines. These results suggest that the evolution of artificial intelligence toward self-maintaining systems may benefit from reasoning not only about hidden states, but also about the evolving validity of what can be known.

Ultimately, the transition from world models to epistemic memory represents a shift from modeling the world to modeling the relationship between the agent and the world. This shift may be important for building intelligent systems that can maintain knowledge validity across changing observation conditions and operate robustly in dynamic environments.

## Acknowledgements

The authors would like to acknowledge the academic discussions and institutional support that contributed to the development of this work. No external funding was received for this research.